\documentclass[letterpaper]{article} 
\usepackage{aaai25}  
\usepackage{times}  
\usepackage{helvet}  
\usepackage{courier}  
\usepackage[hyphens]{url}  
\usepackage{graphicx} 
\usepackage{natbib}  
\usepackage{caption} 
\usepackage{algorithm}
\usepackage{algpseudocode}

\usepackage{newfloat}
\usepackage{listings}
\DeclareCaptionStyle{ruled}{labelfont=normalfont,labelsep=colon,strut=off} 
\floatstyle{ruled}
\newfloat{listing}{tb}{lst}{}
\floatname{listing}{Listing}
\usepackage{amsmath}
\usepackage{array}
\usepackage{subcaption}

\newcolumntype{M}[1]{>{\centering\arraybackslash}m{#1}}

\newtheorem{thm}{Theorem}

\newtheorem{assm}{Assumption}

\newtheorem{lemm}{Lemma}

\newtheorem{claim}{Claim}

\usepackage{todonotes}

\title{FedGAT: A Privacy-Preserving Federated Approximation Algorithm \\ for Graph Attention Networks}
\author {
    Siddharth Ambekar,
    Yuhang Yao,
    Ryan Li,
    Carlee Joe-Wong
}
\affiliations {
    Carnegie Mellon University\\
    \{sambekar,yuhangya,yuanli3,cjoewong\}@andrew.cmu.edu
}

\begin{document}

\maketitle

\begin{abstract}
Federated training methods have gained popularity for graph learning with applications including friendship graphs of social media sites and customer-merchant interaction graphs of huge online marketplaces. However, privacy regulations often require locally generated data to be stored on local clients. The graph is then naturally partitioned across clients, with no client permitted access to information stored on another. Cross-client edges arise naturally in such cases and present an interesting challenge to federated training methods, as training a graph model at one client requires feature information of nodes on the other end of cross-client edges. Attempting to retain such edges often incurs significant communication overhead, and dropping them altogether reduces model performance. In simpler models such as Graph Convolutional Networks, this can be fixed by communicating a limited amount of feature information across clients before training, but GATs (Graph Attention Networks) require additional information that cannot be pre-communicated, as it changes from training round to round. We introduce the Federated Graph Attention Network (FedGAT) algorithm for semi-supervised node classification, which approximates the behavior of GATs with provable bounds on the approximation error. FedGAT requires only one pre-training communication round, significantly reducing the communication overhead for federated GAT training. We then analyze the error in the approximation and examine the communication overhead and computational complexity of the algorithm. Experiments show that FedGAT achieves nearly the same accuracy as a GAT model in a centralised setting, and its performance is robust to the number of clients as well as data distribution.
\end{abstract}

\section{Introduction}

Graph Attention Network (GAT)~\cite{veličković2018graphattentionnetworks} is a graph learning model that employs a self-attention mechanism to overcome the shortcomings of other graph convolution-based models. It is a popular tool in tasks such as node classification and link prediction~\cite{10.1093/bioinformatics/btac559}, with many applications, especially social networks and recommendation systems~\cite{10.1145/3292500.3330989,8954185,zheng2019gmangraphmultiattentionnetwork}. %
However, in many real applications, the underlying graph is naturally partitioned across multiple clients or too large for a single server to train. Such partitioning is often a consequence of data privacy regulations that require data to be stored where it was generated, for example, the General Data Protection Regulation (GDPR) in Europe and Payment Aggregators and Payment Gateways (PAPG) in India. Thus, traditional centralized training of models is no longer possible.

\begin{figure}[t]
    \centering
    \includegraphics[width = \linewidth]{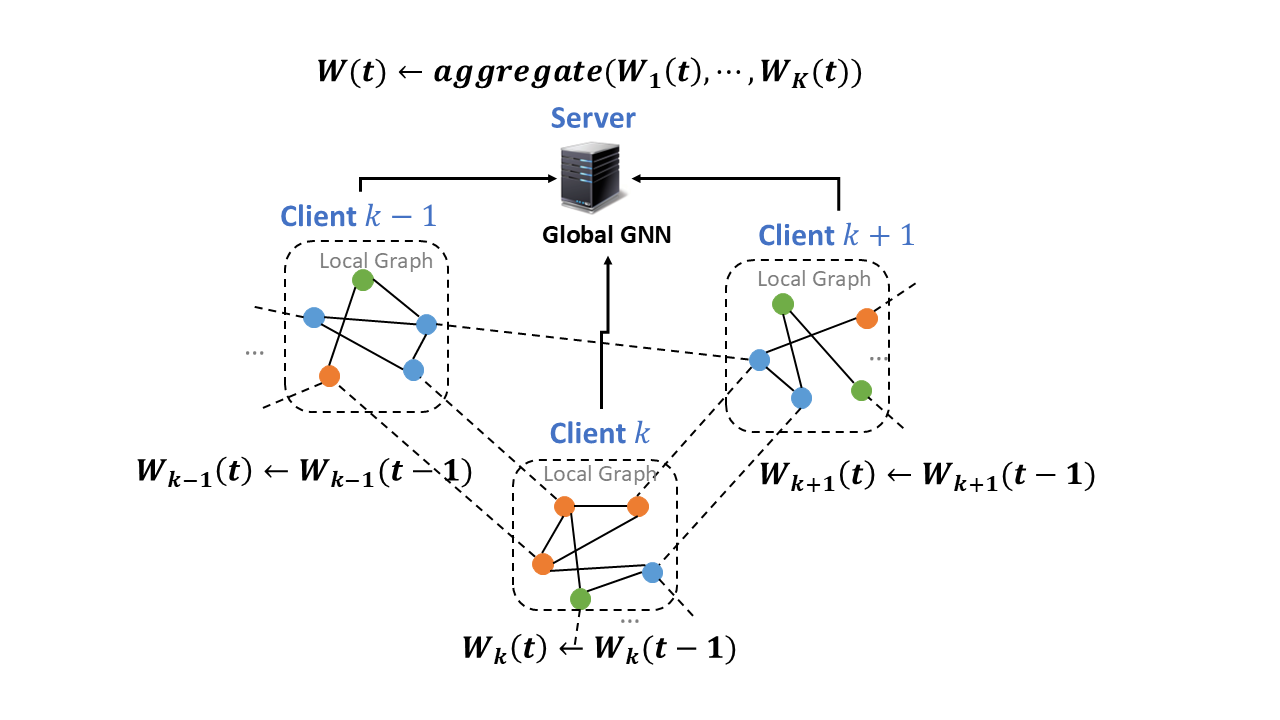}
    \caption{The central model parameters are denoted by $\mathcal{W}$. On receiving a copy of $\mathcal{W}$, clients compute local updates and send updates to the server, where they are aggregated.}
    \label{fig:1}
\end{figure}

Federated learning, which is popular in non-graph settings, somewhat addresses these concerns by keeping local data at each client. Some recent papers have also proposed federated training of graph neural networks~\cite{yao2023fedgcnconvergencecommunicationtradeoffsfederated,he2021fedgraphnnfederatedlearningbenchmark}. Each client trains the model on the local subgraph, occasionally sending its parameters to the central server for aggregation.

We call edges with end-nodes stored in different clients as \emph{cross-client edges}, which arise naturally in partitioned graphs. For example, in social networks such as LinkedIn, many nodes have connections across geographically distinct clients. Both clients that store the end-nodes are aware of the edge. However, due to privacy constraints, it may not be possible for a client to access the node information stored on another (e.g., users in the U.S. to access data on friends in Europe). Cross-client edges thus present a challenge: the node at one end cannot directly access the feature information of the node at the other end, but GATs require this information to compute the attention coefficients.

Recent works ~\cite{yao2023fedgcnconvergencecommunicationtradeoffsfederated,zhang2021subgraphfederatedlearningmissing,chen2021fedgraphfederatedgraphlearning} 
explore federated learning for graph neural networks such as Graph Convolutional Networks (GCN). GATs are much more expressive than GCNs as a result of the attention mechanism, and therefore, can learn from more complex data, outperforming GCNs on several benchmarks~\cite{veličković2018graphattentionnetworks,hu2020heterogeneousgraphtransformer}.
As a consequence of the attention mechanism, though, \textit{sharing the node feature information for cross-client edges is significantly more difficult for GATs than it is for GCNs}. Directly sharing this information introduces significant communication overhead, since the information would have to be shared at every training iteration. Sharing raw node features would also be a violation of privacy. Ignoring cross-client edges solves the privacy and communication concerns, but would lead to loss of information and degrade model performance, as our experiments show (Table~\ref{tab:1}, Figure~\ref{fig:2}). Thus, our goal is to \textit{use a single round of communication for sharing node feature information across clients, ensuring model accuracy while respecting privacy requirements.}

We achieve this goal by using a polynomial approximation of certain terms in the GAT update equation and computing expressions that remain constant throughout the training of the model. By designing these expressions carefully, we ensure that one round of pre-training communication of these expressions allows GAT updates to be computed in a federated manner without further sharing of feature information, thus reducing communication needs. The expressions also do not reveal individual node features, ensuring privacy, which can be further protected by using Homomorphic Encryption for the pre-training communication and computations.
Our work makes the following \textbf{contributions}:

\begin{itemize}
    \item We present \textbf{FedGAT}, an algorithm that enables approximate and privacy-preserving federated training of a Graph Attention Network for node classification. FedGAT requires only one round of pre-training communication that communicates all the information necessary to the GAT. 
    To the best of our knowledge, \textbf{this is the first work that enables the federated training of GATs that computes the GAT updates in a near-exact manner}.
    \item We provide \textbf{bounds on the approximation error of FedGAT} and show that it can be made arbitrarily small. We also analyze FedGAT's computational complexity, privacy guarantees and communication overhead.
    \item We perform \textbf{experiments} on widely used datasets in i.i.d. and non-i.i.d. settings, and demonstrate that a model trained using FedGAT achieves nearly the same performance as a GAT trained in a centralised setting.
\end{itemize}

We give an overview of related work in Section~\ref{sec:related}, before giving the necessary background on GATs in Section~\ref{sec:prelim}. We then introduce FedGAT in Section~\ref{sec:fedgat} and evaluate its performance both analytically (Section~\ref{sec:analysis}) and empirically (Section~\ref{sec:exp}) before concluding in Section~\ref{sec:conclusion}. All proofs are deferred to the appendix due to lack of space.

\section{Related Work}\label{sec:related}

\textbf{Federated learning} has become a common framework for training machine learning models, in which 
a set of clients performs local updates by gradient descent using their local data and shares the optimisation parameters with a coordinating server to keep the local client data secure. FedAvg~\cite{mcmahan2023communicationefficientlearningdeepnetworks} averages the local client updates at every communication round but may not converge when the underlying data distribution is non-iid. 
More federated learning algorithms are proposed to overcome non-iid issue (e.g.FedProx~\cite{li2020federatedoptimizationheterogeneousnetworks},  FedAdam~\cite{reddi2020adaptive}, and APFL~\cite{deng2020adaptive}).

Federated learning is increasingly being applied to \textbf{graph neural networks} (GNNs)~\cite{he2021fedgraphnnfederatedlearningbenchmark}. GATs~\cite{veličković2018graphattentionnetworks}, GCNs~\cite{kipf2017semisupervisedclassificationgraphconvolutional}, and GraphSage~\cite{hamilton2018inductiverepresentationlearninglarge} are examples of popular GNNs. 
Some works on GNNs in the federated setting~\cite{zhang2021subgraphfederatedlearningmissing,li2019deepgcnsgcnsdeepcnns} assume several disjoint graphs distributed across clients. We, instead, consider a single graph partitioned across clients, and the goal is to train a model on the graph using federated learning. Much prior work in our setting ignores cross-client edges~\cite{he2021fedgraphnnfederatedlearningbenchmark,zheng2020asfgnnautomatedseparatedfederatedgraph} to avoid sharing features between clients, with loss of information.

Other works that communicate GNN information across clients include~\cite{Scardapane_2021}, where cross-client features and intermediate results of the GNN are shared across the clients, incurring high communication costs. FedSage+~\cite{zhang2021subgraphfederatedlearningmissing} drops cross-client edges but recovers the missing node information by using a model to learn this information. This method may not fully recover the lost information.~\cite{wan2022bnsgcnefficientfullgraphtraining} suggests randomly sampling cross-client edges and revealing node information, but this violates privacy concerns.
FedGCN~\cite{yao2023fedgcnconvergencecommunicationtradeoffsfederated} instead communicates all cross-client information exactly in a privacy-preserving pre-training round, achieving high accuracy on several benchmarks. However, its communicated information is insufficient for GATs to be trained.

FedGAT uses a single pre-training communication round without dropping cross-client edges but communicates different, carefully designed information to successfully train a GAT model.
To achieve this goal, we make use of
\textbf{Chebyshev polynomials}, which are widely used in function approximation~\cite{ffce8dff-1406-30f4-8ab7-5982dc9e3e4e,descloux1963approximations,trefethen1981rational}. FedGAT uses the Chebyshev series to approximate the GAT update equation with quantifiable error bounds.

\section{Preliminaries}\label{sec:prelim}
\subsection{Graph Attention Networks for Node Classification}
Semi-supervised node classification is an important task in analysing networks that is widely applied to social network analysis (e.g. Cora, Ogbn-Arxiv~\cite{hu2021opengraphbenchmarkdatasets}) and can also be used to learn embeddings for nodes in a graph that are used for downstream tasks such as link prediction and clustering. We focus on designing a federated algorithm that trains GATs for node classification, which is one of the simplest and most important tasks in graph machine learning.

Consider a graph $\mathcal{G} = (\mathcal{V}, \mathcal{E})$, with the node set $\mathcal{V} = [N]$ and edge-set $\mathcal{E}$.  Each node $i$ in the graph has an associated feature vector $x_{i} \in\mathbf{R}^{d}$. We are provided with a subset $\mathcal{V}^{train}\subset \mathcal{V}$, where each node $i\in\mathcal{V}^{train}$ has a label $y_{i}$ associated with it. The task of node classification is to assign labels to nodes to all nodes in $\mathcal{V}\backslash\mathcal{V}^{train}$. 

GATs~\cite{veličković2018graphattentionnetworks} are similar to GCNs~\cite{kipf2017semisupervisedclassificationgraphconvolutional}, but instead of performing a simple graph convolution, GATs perform a weighted convolution, where the weights are calculated using an attention mechanism. Much like GCN layers, each GAT layer constructs embeddings for nodes. Let $h^{(l)}_{i}$ be the $l^{th}$-layer embedding of node $i$. Then, the GAT update equation for generating the embeddings for the $(l+1)$-th layer is
\small\begin{equation}
    h_{i}^{(l + 1)} = \phi\bigg(\sum_{j\in\mathcal{N}_{i}}\alpha^{(l)}_{ij}W^{(l)}h_{j}^{(l)}\bigg)\label{eq:1}
\end{equation}
\normalsize
\small\begin{equation}
    \alpha^{(l)}_{ij} = \displaystyle\frac{e^{(l)}_{ij}}{\sum_{k\in\mathcal{N}_{i}}e^{(l)}_{ik}},\label{eq:2}
\end{equation}
\normalsize
and the attention score between nodes $i$ and $j$ is
\small
\begin{equation}
    e^{(l)}_{ij} = \exp\bigg(\psi\bigg(a_{1}^{(l)T}W^{(l)}h_{i}^{(l)} + a_{2}^{(l)T}W^{(l)}h_{j}^{(l)}\bigg)\bigg),\label{eq:3}
\end{equation}
\normalsize
where $\mathcal{N}_{i}$ denotes the set of one-hop neighbour nodes of node $i$. $W^{(l)}$ is a learnable weight matrix, $a^{(l)}_{1}, a_{2}^{(l)}$ are learnable attention parameters, and $\phi(\cdot), \psi(\cdot)$ are activation functions. For an $L$-layer GAT, the embeddings of each node depend on nodes in its $L$-hop neighborhood.

The attention coefficient $\alpha_{ij}$ between nodes $i$ and $j$ depends on the feature vectors of both nodes. It should be noted that the parameters $a_{1}^{(l)}, a_{2}^{(l)}$ are learnable, and hence, the attention coefficient changes with every training iteration.

\subsection{Graph Attention Networks in the Federated Setting}

%
As before, we are given a graph $\mathcal{G} = (\mathcal{V}, \mathcal{E})$. In the federated setting, there is a central server $\mathcal{S}$, and $K$ clients, where the $k^{th}$ client is denoted by $\mathcal{C}_{k}$. The graph is partitioned across these $K$ clients, such that each client stores a sub-graph $\mathcal{G}_{k} = (\mathcal{V}_{k}, \mathcal{E}_{k})$ of $\mathcal{G}$. A cross-client edge is an edge between nodes stored on two different clients. Each client is aware of its cross-client edges and has a set of training nodes $\mathcal{V}^{train}_{k}\subset\mathcal{V}_{k}$ for which the labels are known. The goal of node classification in the federated setting is to predict labels for nodes in $\mathcal{V}_{k}\backslash\mathcal{V}_{k}^{train}$ for each client $\mathcal{C}_{k}$.

As previously mentioned, the node embeddings of an $L$-layer GAT depend on the feature vectors of nodes in the $L$-hop neighborhood of a node. However, in the federated setting, the nodes in the $L$-hop neighborhood may lie in different clients. Thus, node features need to be shared across clients at each iteration to train a GAT. Sharing raw features may violate certain privacy requirements, so practical federated GAT training requires obfuscating node features.

\section{FedGAT Algorithm}\label{sec:fedgat}

Here, we present the main FedGAT algorithm; we first demonstrate its functioning for a single-layer GAT and then extend it to multi-layer GATs.
We aim to achieve the following goals in addition to training the GAT model:

\begin{itemize}
    \item Nodes in different clients should not be able to retrieve feature vectors of each other. This is to preserve privacy.
    \item There should be only one round of communication that involves the exchange of feature information. This is to minimize communication costs.
    \item Each client recovers an approximate copy of the GAT model in the centralized case, allowing us to match the accuracy of a centrally trained model.
\end{itemize}
The complete FedGAT algorithm is outlined in Appendix A as Algorithm \ref{alg:1} and Algorithm \ref{alg:2}.

\subsection{FedGAT for a Single GAT Layer}

\textbf{GAT updates:} For the first layer of the GAT, the terms $h_{i}^{(0)}$ in \eqref{eq:1} are the feature vectors, and hence, remain fixed throughout the training. However, the attention coefficients $\alpha_{ij}$ must be computed anew at every iteration, as they depend on the learnable attention parameters.
We define 
\small
\begin{equation}
    b_{1} = W^{T}a_{1},\;b_{2} = W^{T}a_{2},\;x_{ij} = b_{1}^{T}h_{i} + b_{2}^{T}h_{j}\label{eq:4}
\end{equation}
\normalsize



Then, we can re-write the coefficients $e_{ij}$ as $e_{ij} = \exp(\psi(x_{ij}))$.
We now assume that $\psi(\cdot)$ is a continuous function, as is typically the case in practice. 
Let $\{P_{n}(x)\vert x\in\mathbf{R}\}_{n = 0}^{\infty}$ be the set of Chebyshev polynomials of the first kind that form a basis of continuous, real-valued functions. Here, $P_{n}$ is a degree $n$ polynomial. Thus, 
we can express the coefficient $e_{ij}$ in terms of a polynomial series as
\small
\begin{equation}
    e_{ij} = \exp(\psi(x_{ij})) = \sum_{n = 0}^{\infty}c_{n}P_{n}(x_{ij})\label{eq:5}
\end{equation}
\normalsize
However, the full expansion is impossible to store in memory; we must, therefore, use a truncated polynomial series. Suppose, we use a polynomial series of degree $p$ to approximate the coefficient $e_{ij}$. Then,
$e_{ij} \approx \sum_{n = 0}^{p}c_{n}P_{n}(x_{ij})$.
Since $P_{n}(x_{ij})$ is a polynomial in $x_{ij}$, we can re-express the approximation as a power series in terms of $x_{ij}$. This works because we truncate the polynomial series at some finite $p$.
\small
\begin{equation}
    e_{ij} \approx \sum_{n = 0}^{p}q_{n}x_{ij}^{n}\label{eq:6}
\end{equation}
\normalsize
This expression forms the basis of the FedGAT algorithm. We shall now present a way to compute this power series in a federated setting.

\noindent\textbf{Federated update computations:} The goal of the algorithm is to compute the node embeddings $h_{i}$ by approximating the scores $e_{ij}$ in the GAT update equation. First, we re-write the GAT update equation in terms of the attention scores
\small
$h_{i} = \phi\bigg(\frac{\sum_{j\in\mathcal{N}_{i}}e_{ij}Wh_{j}}{\sum_{k\in\mathcal{N}_{i}}e_{ik}}\bigg)$\normalsize


Using a truncated polynomial series of maximum degree $p$, re-expressed as a power series, gives us an approximation to the GAT update equation as follows:

\small
\begin{equation}
    h_{i} \approx \phi\bigg(\frac{W\sum_{n = 0}^{p}q_{n}E_{i}^{(n)}}{\sum_{n = 0}^{p}q_{n}F_{i}^{(n)}}\bigg),\label{eq:7}
\end{equation}
\normalsize
where we define 

\small
\begin{equation}
    E_{i}^{(n)} = \sum_{j\in\mathcal{N}_{i}}x_{ij}^{n}h_{j},\; F_{i}^{(n)} = \sum_{j\in\mathcal{N}_{i}}x_{ij}^{n}\label{eq:8}
\end{equation}
\normalsize




If we can somehow compute $E_{i}(n), F_{i}(n)$, we can compute the approximate GAT update $h_i$ in \eqref{eq:7}.

We now turn our attention to a certain type of matrix that enables us to compute these expressions in a federated setting.
Consider node $i$; let $U_{i} = \{u_{1j}, u_{2j}\}_{j\in\mathcal{N}_{i}}$ 
be a set of orthonormal vectors such that $u_{1j}^{T}u_{1k} = \delta_{jk}$, $u_{2j}^{T}u_{2k} = \delta_{jk}$ and $u_{1j}^{T}u_{2k} = 0$ $\forall j, k\in\mathcal{N}_{i}$, where $\delta_{jk}$ is the Kronecker delta. 
Define $\mathcal{U}_{i} = \{\mathcal{U}_{i}(j)\}_{j\in\mathcal{N}_{i}}$, where

\small
\begin{equation}
    \mathcal{U}_{j} = \frac{1}{2}\bigg(u_{1j}u_{1j}^{T} + u_{2j}u_{2j}^{T} + ru_{1j}u_{2j}^{T} + \frac{1}{r}u_{2j}u_{1j}^{T}\bigg)\label{eq:9}
\end{equation}
\normalsize

Here, $r\in\mathbf{R}$ is a constant chosen at random. \textbf{We note some useful properties of these matrices}: $\forall j, k\in\mathcal{N}_{i}$, $\mathcal{U}_{j}^{2} = \mathcal{U}_{j}$ and $\mathcal{U}_{j}\mathcal{U}_{k} = 0$. We now define

\small
\begin{equation}
    D_{i} = \sum_{j\in\mathcal{N}_{i}}x_{ij}\mathcal{U}_{j}\label{eq:10}
\end{equation}
\normalsize

\small
\begin{equation}
    K_{1i} = \sqrt{2}\sum_{j\in\mathcal{N}_{i}}u_{1j},\; K_{2i} = \sqrt{2}\sum_{j\in\mathcal{N}_{i}}u_{1j}h_{j}^{T}\label{eq:11}    
\end{equation}
\normalsize

where $\mathcal{U}_{j}$ is defined in \eqref{eq:9}. Observe that $D_{i}^{n} = \sum_{j\in\mathcal{N}_{i}}x_{ij}^{n}\mathcal{U}_{j}$. We can compute $E_{i}(n), F_{i}(n)$, as defined in \eqref{eq:8}, as follows:

\small
\begin{equation}
    E_{i}^{(n)} = \bigg(K_{1i}^{T}D_{i}^{n}K_{2i}\bigg)^{T},\; F_{i}^{(n)} = K_{1i}^{T}D_{i}^{n}K_{1i}
    \label{eq:12}    
\end{equation}
\normalsize


Now, we move to the final step in describing FedGAT. Consider the expression for $D_{i}$ as described in \eqref{eq:10} and $x_{ij}$ \eqref{eq:4}. Suppose the features $h_{i}, h_{j}\in\mathbf{R}^{d}$. Then, $x_{ij} = \sum_{s = 1}^{d}b_{1}(s)h_{i}(s) + b_{2}(s)h_{j}(s)$, where $h_{i}(s)$ denotes the $s^{th}$ component of the vector $h_{i}$. Then, $D_{i} = \sum_{j\in\mathcal{N}_{i}}\mathcal{U}_{j}\sum_{s = 1}^{d}b_{1}(s)h_{i}(s) + b_{2}(s) + h_{j}(s)$. Changing the order of summation and defining



\small
\begin{equation}
    M_{1i}(s) = \sum_{j\in\mathcal{N}_{i}}h_{i}(s)\mathcal{U}_{j},\; M_{2i}(s) = \sum_{j\in\mathcal{N}_{j}}h_{j}(s)\mathcal{U}_{j},\label{eq:13}
\end{equation}
\normalsize

we can express $D_{i}$ in terms of these matrices as

\small
\begin{equation}
    D_{i} = \sum_{s = 1}^{d}b_{1}(s)M_{1i}(s) + b_{2}(s)M_{2i}(s)\label{eq:14}
\end{equation}
\normalsize

Knowing $M_{1i}(s), M_{2i}(s), K_{1i}, K_{2i}$ then suffices for us to compute an approximation of $e_{ij}$ in \eqref{eq:6}.

\noindent\textbf{Pre-training communication:} Since the matrices $M_{1i}(s), M_{2i}(s), K_{1i}, K_{2i}$ are independent of the learnable parameters of the GAT, they remain constant throughout the training phase. Computing and sharing these parameters thus constitutes FedGAT's single pre-training communication round. It is important to note that the matrices $M_{1i}(s), M_{2i}(s), K_{1i}, K_{2i}$ \textbf{do not reveal individual node features}, ensuring privacy requirements are satisfied.

The \textbf{FedGAT algorithm} then runs as follows:

\textbf{Step 1: } The central server receives the feature vectors of all the nodes in the graph.

\textbf{Step 2: Pre-training.} $\forall i\in\mathcal{V}$, the server computes $\{M_{1i}(s), M_{2i}(s)\}_{s = 1}^{d}, K_{1i}, K_{2i}$ as defined in \eqref{eq:13} and \eqref{eq:11}, and shares them with the appropriate clients. 

\textbf{Step 3: Training rounds.} 
At each round, clients perform approximate GAT updates using \eqref{eq:7},\eqref{eq:12},\eqref{eq:14}. A suitable parameter aggregation scheme is used to update global parameters, which are sent back to the clients for the next round.

\subsection{FedGAT for Multiple GAT Layers}
To extend the algorithm to multiple layers, we assume that nodes are permitted to view the embeddings of any other node generated after the first GAT layer, including nodes on other clients. This assumption can be justified by the fact that the GAT update is highly non-linear in the input vectors and thus difficult to invert, so do not reveal private information.

The first layer embeddings, which require the raw features, are computed using the single layer FedGAT algorithm. Once these embeddings are available, for layers $l > 1$, we can use the regular GAT update \eqref{eq:1}.






\subsection{Model Training and Parameter Aggregation}

After the pre-training communication, FedGAT alternates between local training and parameter aggregation for a number of discrete rounds $t = 1,2,\ldots, T$, as is typical in federated learning. 
In each global training round, clients send their local updates to a central server for aggregation. To aggregate client parameters, we have used \textbf{FedAvg} \cite{mcmahan2023communicationefficientlearningdeepnetworks}. However, FedGAT may use any Federated Learning algorithm for parameter aggregation, including, but not limited to, \textbf{FedAvg}\cite{mcmahan2023communicationefficientlearningdeepnetworks}, \textbf{FedProx}\cite{li2020federatedoptimizationheterogeneousnetworks}, \textbf{ADMM}~\cite{Glowinski1975, GABAY197617}, etc.

\section{Analysis of FedGAT}\label{sec:analysis}

In this section, we analyze the computation, communication overhead and approximation error of FedGAT.

The graph is denoted by $\mathcal{G} = (\mathcal{V} = [N], \mathcal{E})$. There are $K$ clients $\{\mathcal{C}_{k}\}_{k = 1}^{K}$. The maximum node degree is denoted by $B$. The feature vectors $h_{i}$ are at most $d$-dimensional. The GAT used is $L$-layers deep, and the size of the largest $L$-hop neighborhood of the sub-graph of any client is $B_{L}$. The $m$-hop neighborhood of a sub-graph is the union of the $m$-hop neighborhoods of all nodes in the sub-graph. A degree $p$ approximation is used for computing attention coefficients. The loss function for the model is $\mathcal{L}(\mathcal{W})$, and the model parameters are denoted by $\mathcal{W}$.

\begin{assm}[$\beta$-smoothness and twice differentiable]
    The loss function for each client $\mathcal{C}_{k}$, $\mathcal{L}_{k}(\mathcal{W}_{k})$, is twice differentiable and $\beta$-smooth with respect to the model parameters $\mathcal{W}_{k}$ where $\beta>0$.
    \small$$\|\nabla\mathcal{L}_{k}(\mathcal{W}_{k}) - \nabla\mathcal{L}_{k}(\mathcal{W}_{k}')\|_{2} \leq \beta\|\mathcal{W}_{k} - \mathcal{W}_{k}'\|_{2}$$\normalsize
    \label{assm:1}
\end{assm}

\begin{assm}[Bounded norm of model parameters]
    For an $L$-layer GAT, with model parameters, let the $l^{th}$ layer have $\Omega(l)$ attention heads. The parameters of the $\omega^{th}$ attention head are the attention parameters $a_{1}^{(l)}(\omega), a_{2}^{(l)}(\omega)$, and the weight matrix $W^{(l)}(\omega)$. Then, $\forall l\leq L$ and $\forall \omega\leq\Omega(l)$
    \begin{center}
    \begin{tabular}{ccc}
       \small$\|a_{1}^{(l)}(\omega)\|_{2}\leq 1$\normalsize  & \small$\|a_{2}^{(l)}(\omega)\|_{2}\leq 1$\normalsize & \small$\|W^{(l)}(\omega)\|_{2}\leq 1$\normalsize \\ 
    \end{tabular}
    \end{center}
    \label{assm:2}
\end{assm}

\begin{assm}[Bounded norm of node embeddings]
    All node features have at most unit norm in the $\mathbf{L}_{2}$-norm.\label{assm:3}
\end{assm}

\begin{assm}
    All the activation functions used in the GAT model are Lipschitz continuous and monotone.
    \label{assm:4}
\end{assm}

Assumption~\eqref{assm:1} is standard in the Federated Learning literature. While Assumption~\eqref{assm:2} is not strictly true in practical models, choice of parameter initialisation, learning rate, as well as regularisation can ensure that it is loosely satisfied. Assumption~\eqref{assm:3} can be easily ensured by appropriate scaling of the features, without affecting model performance. Finally, some of the proofs for error bounds require Assumption~\eqref{assm:4}. Since ReLU, LeakyReLU activations are widely used, and satisfy this assumption, we feel it is reasonable.

\subsection{Communication Overhead}

The communication between server and clients is of two kinds: the pre-training communication, and the parameter aggregation during training. We shall only deal with the pre-training communication overhead, as the overhead of parameter aggregation is equivalent to that of the parameter aggregation procedure chosen for standard federated learning.

The pre-training communication involves two steps - in the first step, each client uploads node features of all nodes it has access to. The server then computes the matrices $\{M_{1i}(s), M_{2i}(s)\}_{s = 1}^{d}, K_{1i}, K_{2i}$ and then transmits them to the appropriate nodes. Then, the cost of the upload step from clients to servers is simply $\mathcal{O}(Nd)$.

\begin{thm}[FedGAT overhead]
    For an $L$-layer GAT, the cost at the server of computing the pre-training communication values is $\mathcal{O}\left( KB_{L}d\left( B^{2} + B^{3} \right) \right)$. The resulting communication overhead is $\mathcal{O}\left( KB_{L}dB^{3} \right)$.\label{thm:1}
\end{thm}
The $B^{3}$ dependence on the maximum node degree places limits on the scalability of FedGAT to dense graphs. However, for sparse graphs, the algorithm has \emph{linear runtime} in the number of nodes in the graph. In Appendix F, we introduce \textbf{Vector FedGAT} that reduces this complexity bound.

\subsection{Approximation Error Bounds for FedGAT}

The FedGAT algorithm computes an approximation to the attention coefficients $\alpha_{ij}$ via a Chebyshev series. Here, we present some results on the error in the approximation, and how it propagates across multiple GAT layers.

The following theorem \cite{doi:10.1137/1.9781611975949} quantifies the error in approximation using a truncated Chebyshev series.

\begin{thm}[Chebyshev approximation]

Let $f:[-1, 1]\rightarrow\mathbf{R}$ be a $k$ times differentiable function, with each of $f^{(1)}, f^{(2)}, ..., f^{(k - 1)}$ being absolutely continuous, and $f^{(k)}$ having bounded variation $V$. The series $s_{p}(f; x) = \sum_{i = 0}^{p}a_{i}T_{i}(x)$ is defined as the truncated Chebyshev series of $f$
where $T_{i}(x)$ is the $i^{th}$ Chebyshev polynomial of the first kind.

    Then, $\forall p>k$, the error in approximation of $f$ with $s_{n}(f; x)$ satisfies the relation

    $$\|s_{p}(f; x) - f\|_{\infty} \leq \frac{2V}{\pi k(p - k)^{k}}$$\label{thm:2}
    
\end{thm}
Thus, the Chebyshev approximation can be made arbitrarily precise by choosing an appropriate expansion degree $p$.


We now use this result to analyze the \textbf{error propagation across layers} in the FedGAT model.

In the FedGAT algorithm for a single layer GAT, let the true attention coefficients be $\alpha_{ij}$, and the approximate coefficients be $\hat{\alpha}_{ij}$. Similarly, let $e_{ij}$, as defined in \eqref{eq:3}, be the true attention scores, and $\hat{e}_{ij}$ be the approximate score.

\begin{thm}
    [Error in attention coefficients]
    
    In the single layer FedGAT algorithm, let $\|\hat{e}_{ij} - e_{ij}\|_{2} = \epsilon$. Then, $\|\hat{\alpha}_{ij} - \alpha_{ij} \|_{2} \leq \alpha_{ij}\frac{2\epsilon}{1 - \epsilon}$\normalsize.
    
    Thus, as the Chebyshev approximation error $\epsilon$ in the attention scores $e_{ij}$ decreases, so does the error in $\alpha_{ij}$.\label{thm:3}
\end{thm}

\begin{thm}
    [Error in layer 1 embeddings]

    Let $h_{i}^{(1)}$ be the embedding of node $i$ obtained after the first GAT layer and $\hat{h}_{i}^{(1)}$ be the embedding generated by the FedGAT layer. 
    
    If $\|\hat{e}_{ij} - e_{ij}\|\leq \epsilon$, then,

    \small$$\|h_{i}^{(1)} - \hat{h}_{i}^{(1)}\|\leq \frac{2\kappa_{\phi}\epsilon}{1 - \epsilon}$$\normalsize

    \noindent where $\kappa_{\phi}$ is the Lipschitz constant of the activation $\phi(\cdot)$.\label{thm:4}
\end{thm}

\begin{thm}
    [Error propagation across layers]

    Let $h_{i}^{(l)}$ be the embeddings of node $i$ obtained after the $l^{th}$ GAT layer. Let $\hat{h}_{i}^{(l)}$ be the $l^{th}$ layer embedding in FedGAT algorithm.

    Let $\|h_{i}^{(l - 1)} - \hat{h}_{i}^{(l - 1)}\|\leq \delta$, and let $\delta\leq \frac{\log(c)}{2\kappa_{\psi}}$, where $c$ is a positive constant. Then, the following results hold:

    \begin{equation}
        \epsilon  = \|\hat{e}_{ij}^{(l)} - e_{ij}^{(l)}\| \leq 2c\kappa_{\psi}\delta\label{eq:15}
    \end{equation}

    \begin{equation}
        \|\hat{\alpha}_{ij}^{(l)} - \alpha_{ij}^{(l)}\| \leq \alpha_{ij}^{(l)}\frac{2\epsilon}{1 - \epsilon} = \alpha_{ij}^{(l)}\varepsilon\label{eq:16}
    \end{equation}

    \begin{equation}
        \|\hat{h}_{i}^{(l)} - h_{i}^{(l)}\|\leq \kappa_{\phi}(\varepsilon + \delta)\label{eq:17}
    \end{equation}\label{thm:5}\normalsize
\end{thm}

From Theorem \eqref{thm:5}, we see that the error in layer $l - 1$ scales by a factor of $\kappa_{\psi}$ in the next layer. Thus, in an $L$-layer GAT, an error of $e$ in the first FedGAT layer becomes an error of $\mathcal{O}(\kappa_{\psi}^Le)$ in the final node embeddings. Thus, an appropriate degree of approximation $p$ is necessary to ensure that the initial error is small enough to remain small after passing through the GAT layers. Since most tasks require less than 3 GAT layers, and Theorem \eqref{thm:2} establishes good convergence rates for Chebyshev approximation of differentiable functions, the FedGAT approximation scheme is sound.

\subsection{Privacy Analysis of FedGAT}

The FedGAT algorithm uses expressions that reveal at most aggregate feature information of a node neighbourhood, but \emph{never} releases individual node information. In this section, we show that na\"ive methods to reconstruct private node features from the information shared by FedGAT are infeasible due to the aggregation of information across multiple nodes.

\subsubsection*{Client-side privacy}

The information that is shared across clients in the pre-training round is the set of matrices $\left\{M_{1i}(s), M_{2i}(s)\right\}_{s = 1}^{d}$ and $K_{1i}, K_{2i}$ for each node $i\in\mathcal{V}$. We briefly show why these matrices cannot reveal individual node information. The matrix $K_{2i}$, as defined in \eqref{eq:11}, contains a sum of outer products of the vectors $u_{1j}$ and the node feature vectors $h_{j}$ $\forall j\in\mathcal{N}_{i}$. The vector $K_{1i}$, also defined in \eqref{eq:11} is a sum of the orthonormal vectors, and cannot reveal any node feature information.

To an observer at the client, $K_{1i}$ is a single vector. Combining it with $K_{2i}$ via matrix vector product reveals only the \emph{aggregate node feature information}.

\small\[
    K_{1i}^{T}K_{2i} = 2\sum_{j\in\mathcal{N}_{i}}h_{j}^{T}
\]\normalsize

If a given node $i$ has neighbours in other clients, as long as there is more than one such neighbour, no individual node feature information is revealed. \emph{If a node has only one cross-client neighbour, in that case, it becomes necessary to entirely drop that neighbour to prevent information leakage.}

Thus, we have established that $K_{1i}$ and $K_{2i}$ cannot reveal individual node feature information as long as multiple cross-client neighbours exist. 

We now investigate the matrices $M_{1i}(s)$ and $M_{2i}(s)$. Referring to \eqref{eq:11}, \eqref{eq:12} and \eqref{eq:13}, we observe the following - the matrix $M_{1i}(s)$ contains information exclusively about node $i$ itself and therefore, does not present itself as an avenue to reveal node feature information. Also, since nowhere do we reveal the orthonormal vectors $u_{1j}, u_{2j}$ or the matrices $\mathcal{U}_{j}$ (which are constructed from the orthonormal vector set, as defined in \eqref{eq:9}), it is not possible to use $K_{1i}, K_{2i}$ or $M_{1i}(s)$ to recover these quantities. Finally, $M_{2i}(s)$ \eqref{eq:13} is the only other matrix containing any relevant information. However, once again, it is encoded as an aggregate of all neighbour node feature vectors. The only meaningful information we can recover from it is the following - 

\small\[\begin{aligned}
    K_{1i}^{T}M_{2i}(s)K_{2i} = \sum_{j\in\mathcal{N}_{i}}h_{j}(s)
\end{aligned}\]\normalsize

Once again, we only recover information about the aggregate, and never the individual nodes.

\subsubsection*{Server-side privacy}

At the server end, it is necessary to ensure that no individual node features are revealed. This can be achieved by \emph{Homomorphic Encryption}, which enables direct computation on encrypted data. Using this adds a computational overhead that is still linear in the number of node features. We have not implemented the encryption scheme in our algorithm, but it is a potential extension.

Thus, the FedGAT algorithm reveals only \emph{aggregate} node feature information, but not individual node features.

\section{Experimental Results}\label{sec:exp}

\subsection{Experimental Setup}

We perform experiments for a node classification task in a federated setting with Cora, Citeseer, and Pubmed datasets. For all experiments, we have used FedAvg as the parameter aggregation algorithm. We have used a degree 16 Chebyshev approximation to the attention coefficients in FedGAT. All the other model parameters are kept the same as the centralized GAT. Accuracy results are averaged over 10 runs of the model. We also measure the pre-training communication overhead. Additional experiments are in the appendix.

\noindent\textbf{Methods Compared } We measure the performance and other statistics of FedGAT with the following methods -

\begin{itemize}
    \item \textbf{Centralised GAT and GCN} - All the data is stored on a single server, where we train a GCN or GAT. The GAT and GCN hyperparameters are based on \citet{veličković2018graphattentionnetworks} and \citet{kipf2017semisupervisedclassificationgraphconvolutional} respectively. We do not use neighbour sampling/dropout for fair comparison to FedGAT. GAT upper-bounds FedGAT's achievable accuracy, and comparison to GCN quantifies the benefit of using the attention mechanism in FedGAT.
    \item \textbf{Federated GAT with no cross-client communication (DistGAT)} - The graph is partitioned across several clients; however, all cross-client edges are dropped. Thus, each client trains its model independently on its local subgraph, and FedAvg is used for parameter aggregation. This experiment quantifies the benefit of retaining cross-client edges with FedGAT. The setting is equivalent to ~\citet{he2021fedgraphnnfederatedlearningbenchmark,zheng2020asfgnnautomatedseparatedfederatedgraph}. 
    \item \textbf{FedGCN as a baseline -}
    Our method is heavily inspired by FedCGN~\cite{yao2023fedgcnconvergencecommunicationtradeoffsfederated}. As such, we shall compare the performance of FedGAT with FedGCN under a multitude of circumstances (data heterogeneity, different number of clients). 
\end{itemize}

To incorporate the i.i.d and non-i.i.d settings, we have used the Dirichlet label distribution with parameter $\beta$ as outlined in \cite{hsu2019measuringeffectsnonidenticaldata}.
The detailed experiment set-up and extended results are included in the appendix.

\subsection{Model Accuracy}

\begin{figure*}[t]
    \centering
    \begin{subfigure}[b]{0.22\linewidth}
    \centering
        \includegraphics[width = \linewidth]{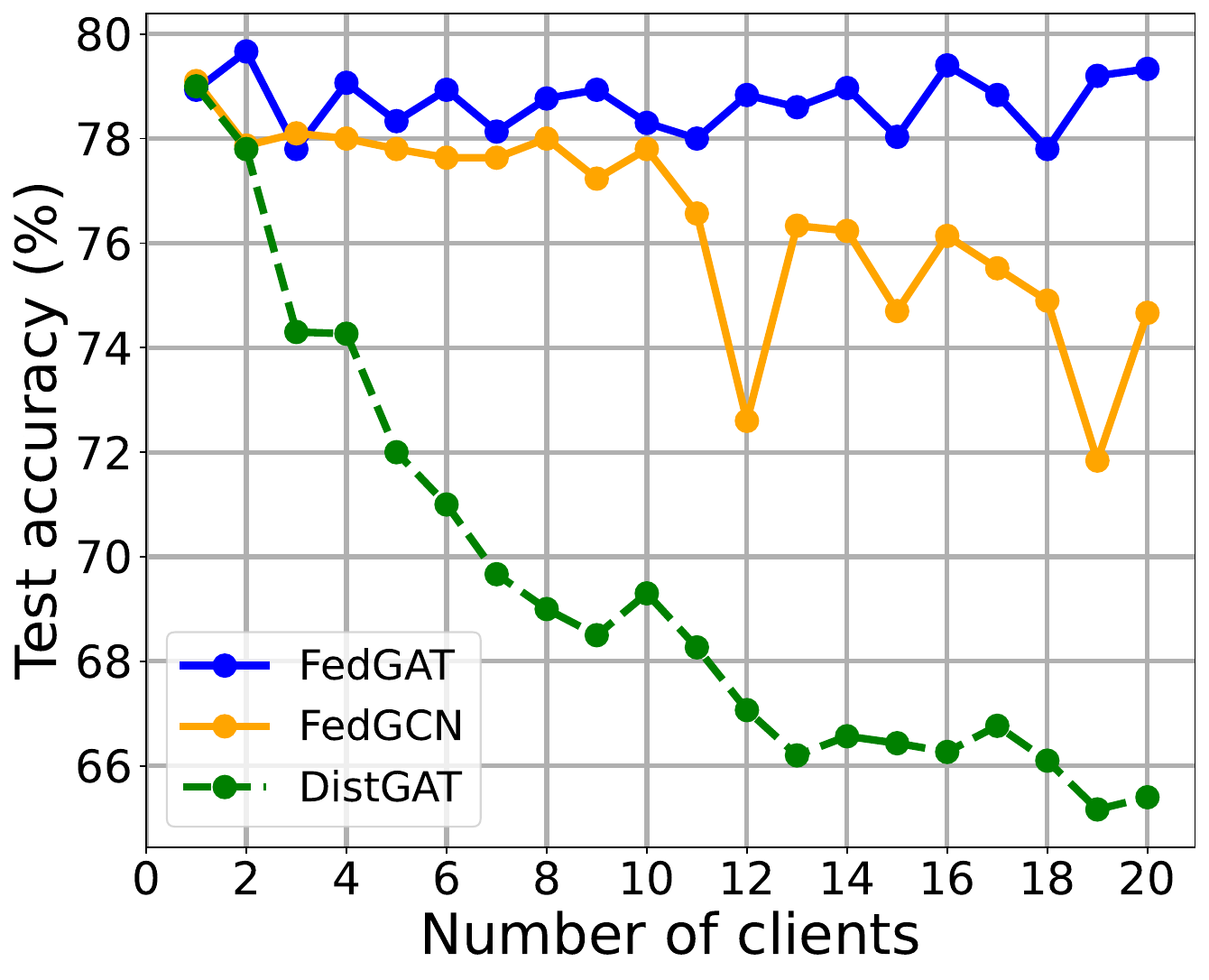}
        \caption{$\beta = 1$ (non-iid)}
    \end{subfigure}
    \begin{subfigure}[b]{0.22\linewidth}
    \centering
        \includegraphics[width = \linewidth]{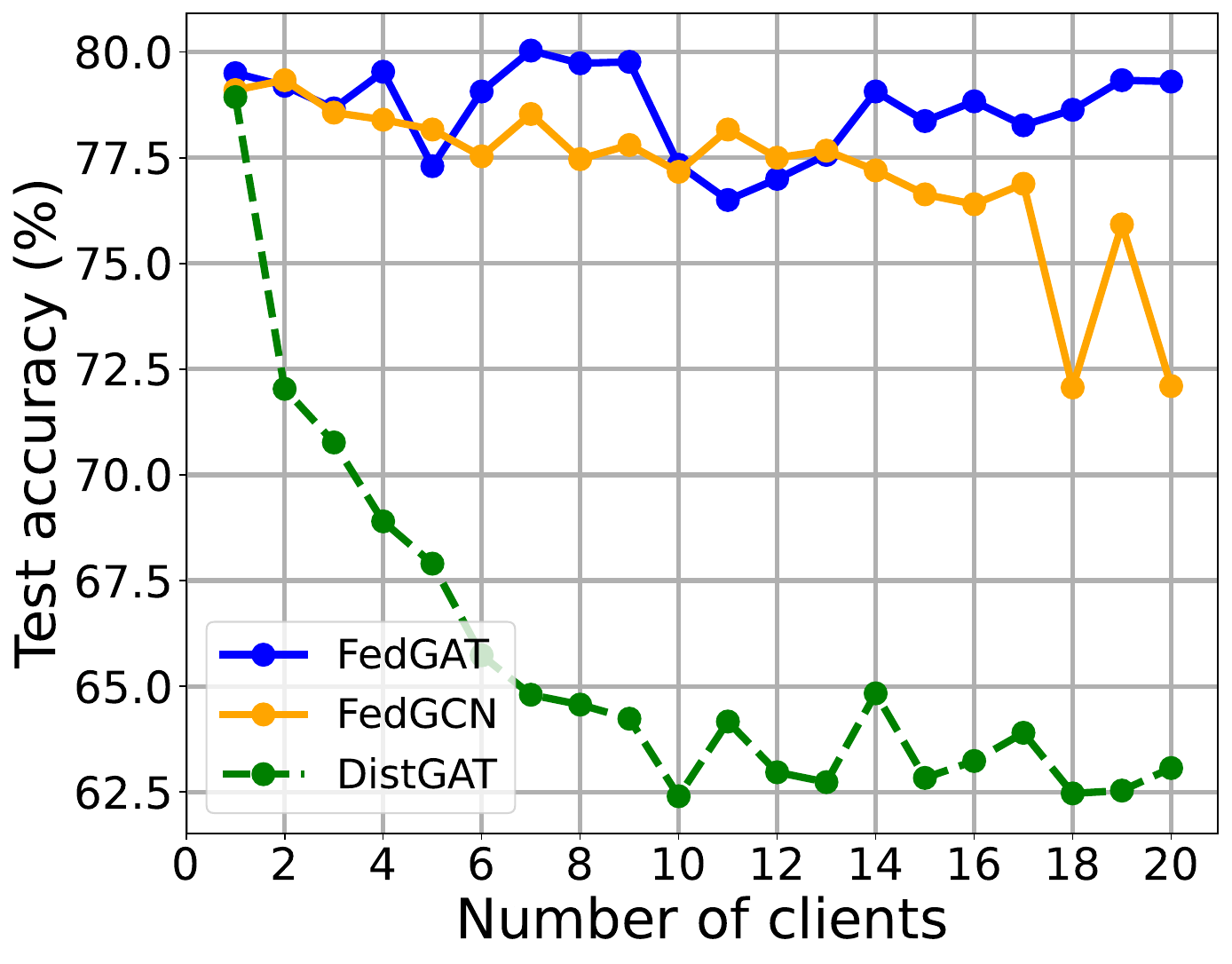}
        \caption{$\beta = 10000$ (iid)}
    \end{subfigure}
    \centering
    \begin{subfigure}[b]{0.22\linewidth}
    \centering
        \includegraphics[width = \linewidth]{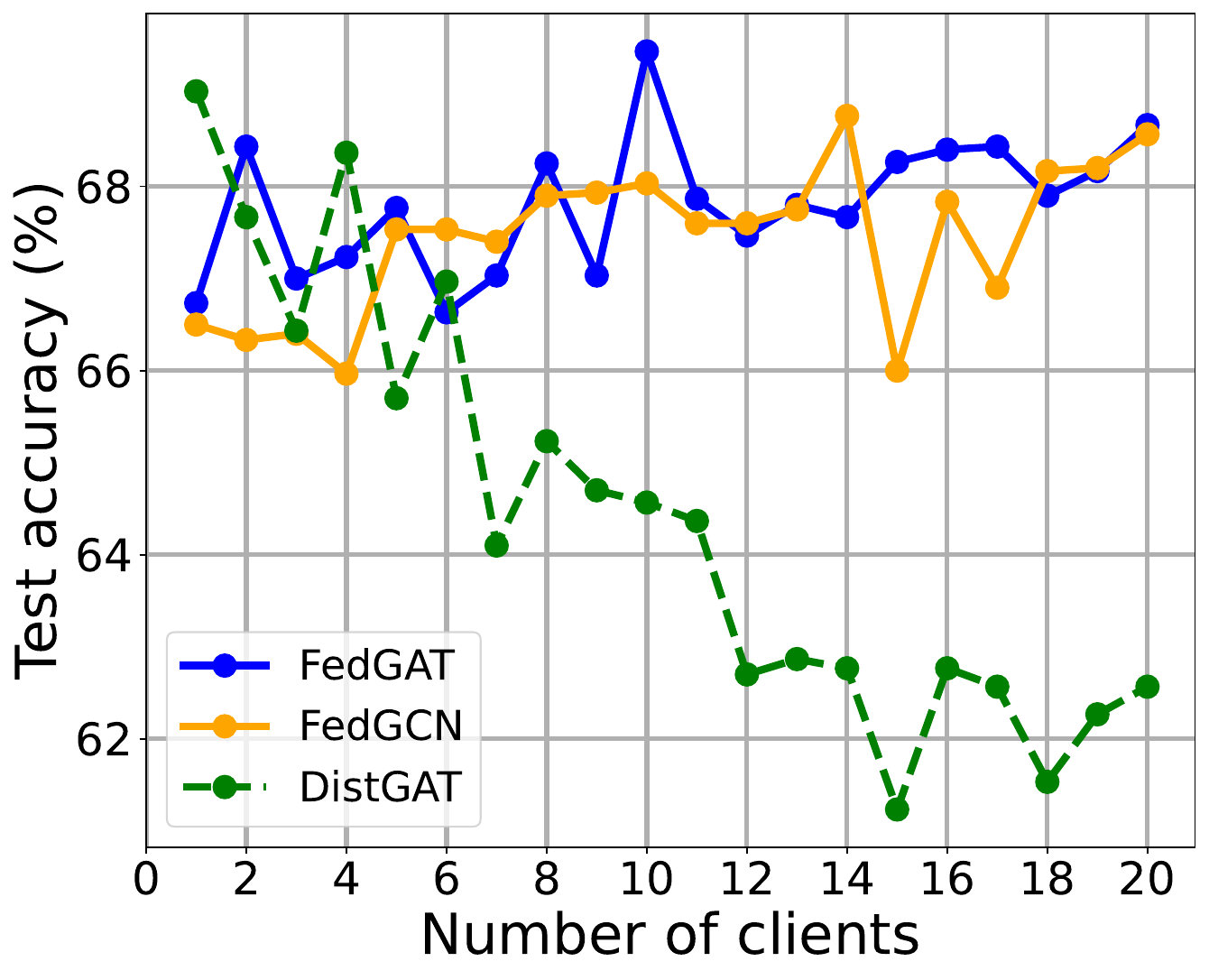}
        \caption{$\beta = 1$ (non-iid)}
    \end{subfigure}
    \begin{subfigure}[b]{0.22\linewidth}
    \centering
        \includegraphics[width = \linewidth]{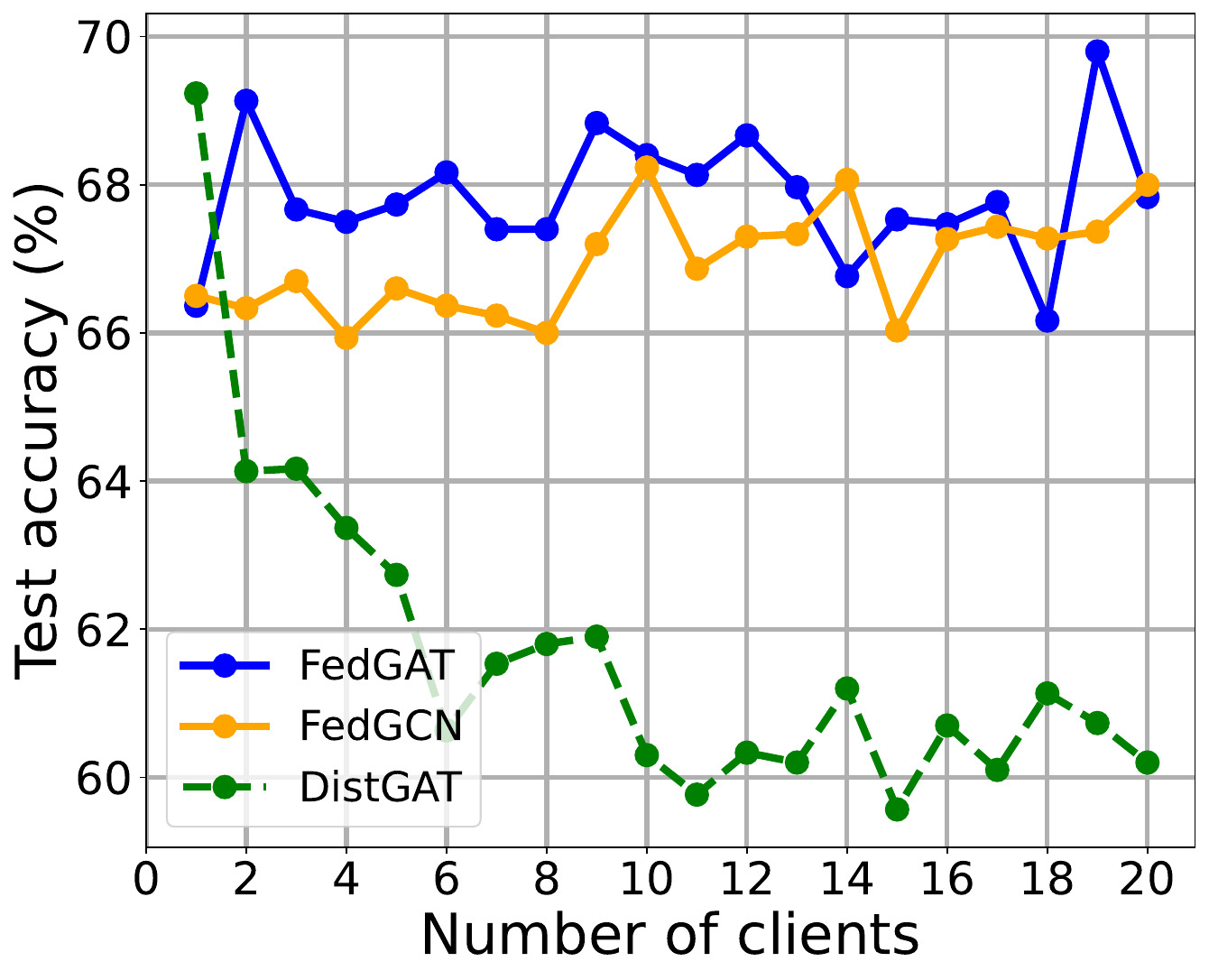}
        \caption{$\beta = 10000$ (iid)}
    \end{subfigure}
    \caption{Test accuracy v/s number of clients for iid and non-iid data distribution on the Cora (a,b) and Citeseer (c,d) dataset. FedGAT outperforms FedGCN and DistGAT.}
    \label{fig:2}
\end{figure*}

We performed experiments with Cora, Citeseer and Pubmed, with number of clients from 1 to 20, and $\beta$ = 1(non-iid), 10000(iid). As can be seen from Figure~\ref{fig:2}, the test accuracy of the FedGAT algorithm is very consistent for a varying number of clients. However, DistGAT suffers a drop in model performance. This is expected since dropping cross-client information leads to a loss of information. This accuracy drop is greater for a larger number of clients since more clients will result in more crossing edges.

\begin{table}[t]
    \centering
    \tabcolsep=0.06cm
    \begin{tabular}{|M{2.42cm}|M{1.32cm}|M{1.32cm}|M{1.32cm}|}
    \hline
      \textbf{Method} & \textbf{Cora}  &  \textbf{Citeseer} & \textbf{Pubmed}\\
       \hline\hline
       GCN  & 0.805 & 0.672 & 0.758 \\
       \hline
       GAT  & 0.813 & 0.725 & 0.795 \\
       \hline
       DistGAT(10 clients,non-iid) & $0.684\pm0.013$ & $0.664\pm0.014$ & $0.769\pm0.01$ \\
       \hline
       DistGAT(10 clients,iid)  & $0.645\pm0.012$ & $0.635\pm0.007$ & $0.745\pm0.0092$ \\
       \hline
       FedGCN(10 clients,non-iid) & $0.778\pm0.003$ & $0.68\pm0.002$ & $0.772\pm0.003$\\
       \hline
       FedGCN(10 clients, iid) & $0.771\pm0.003$ & $0.682\pm0.008$ & $0.774\pm0.002$\\
       \hline
       FedGAT(10 clients,non-iid) & $0.80\pm0.005$ & $0.699\pm0.004$ & $0.789\pm0.004$ \\
       \hline
       FedGAT(10 clients,iid)  & $0.802\pm0.003$ & $0.694\pm0.006$ & $0.787\pm0.005$ \\
       \hline
    \end{tabular}
    \caption{FedGAT test accuracy is quite close to the regular GAT model, but DistGAT suffers from performance loss due to dropping cross-client edges. FedGAT also outperforms FedGCN on all benchmarks. Only peak accuracies are reported for GAT and GCN.}
    \label{tab:1}
\end{table}

\textbf{FedGAT accuracy is unaffected by iid and non-iid data,} as seen from both Figure~\ref{fig:2} and Table~\ref{tab:1}. This is because the algorithm does not drop any cross-client edges. As a result, there is no loss in information, thus ensuring that data heterogeneity does not affect model training. Moreover, \textbf{FedGAT almost achieves the same accuracy as GAT}. It should also be noted that FedGAT outperforms FedGCN and even centralised GCN, both under iid and non-iid data distributions (Table ~\ref{tab:1}, Figure ~\ref{fig:2}). This justifies our motivation for designing a federated version of GATs. 

\subsection{Communication Cost}

We measured the communication cost of the FedGAT algorithm in terms of the number of scalar values it transfers during the pre-training communication round.
\begin{figure}[t]
    \centering
    \begin{subfigure}[b]{0.4\linewidth}
    \centering
        \includegraphics[width = \linewidth]{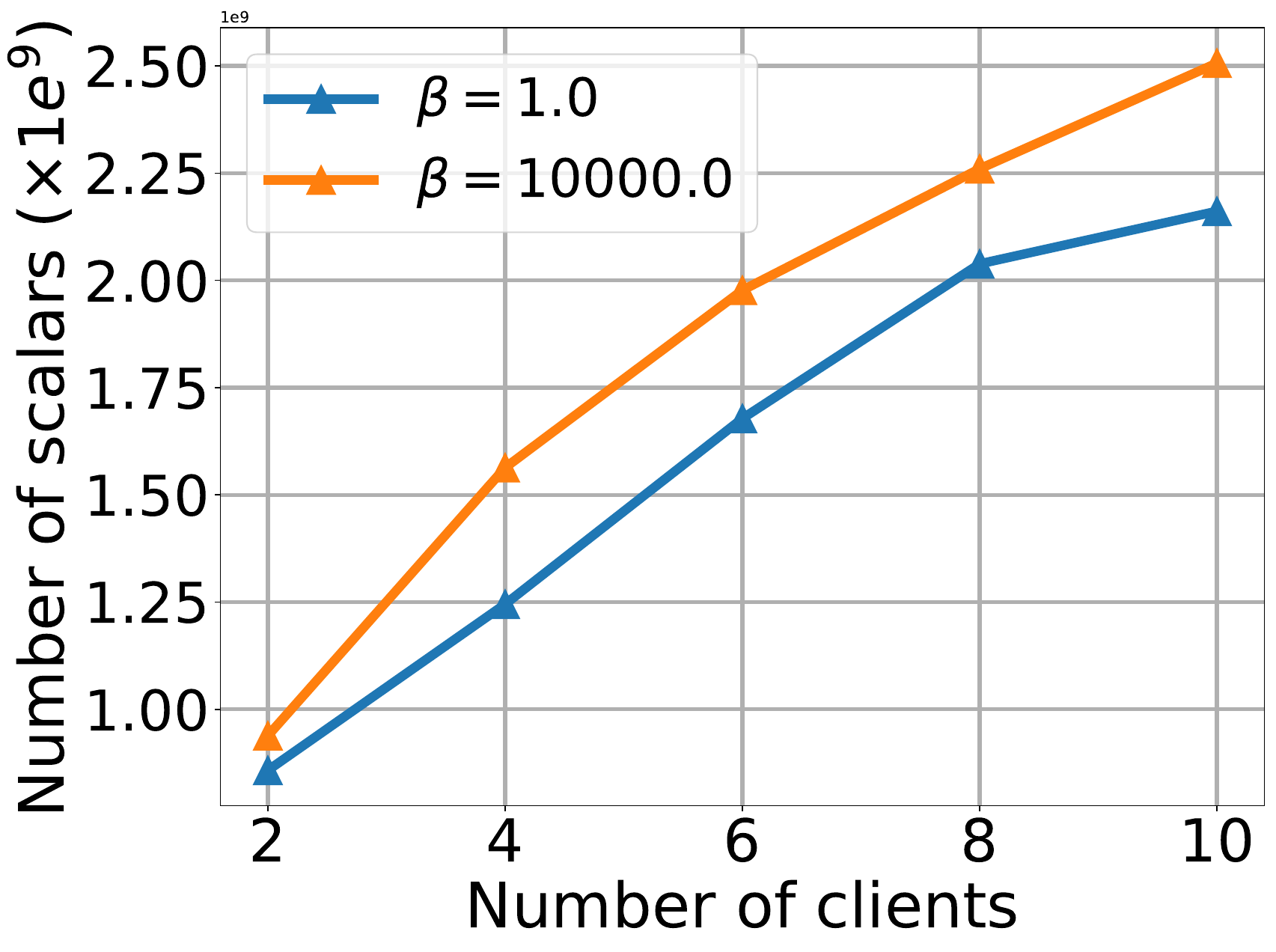}
        \caption{Cora dataset.}
    \end{subfigure}
    \begin{subfigure}[b]{0.4\linewidth}
    \centering
        \includegraphics[width = \linewidth]{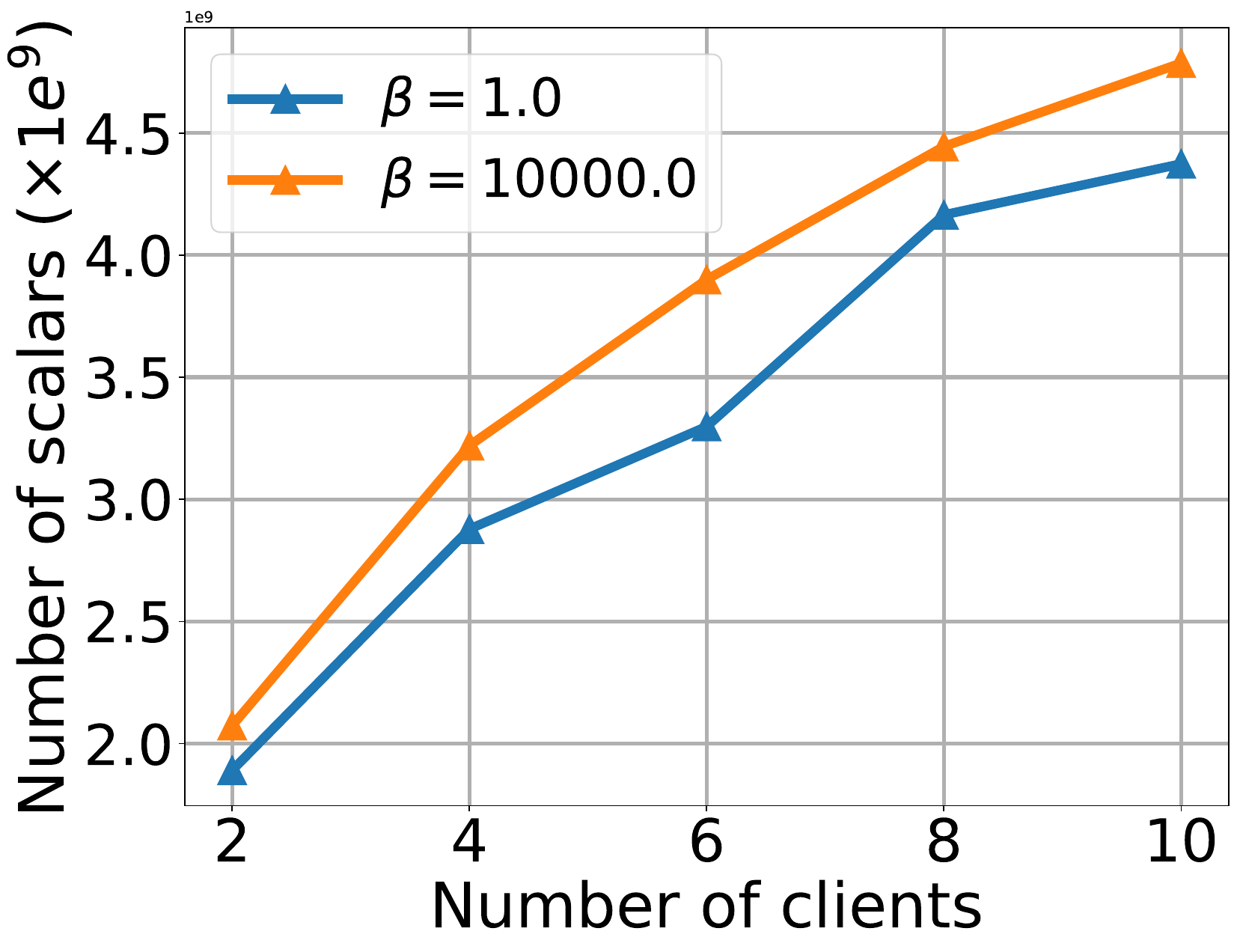}
        \caption{Citeseer dataset.}
    \end{subfigure}
    \caption{The FedGAT pre-training communication increases with clients due to an increase in cross-client edges, leading to a larger sub-graph on each client. The cost is higher for an iid distribution due to increased crossing edges.}
    \label{fig:3}
\end{figure}
The pre-training communication depends on the number of cross-client edges. An iid data distribution mixes up all the nodes across the clients. Assuming that nodes with similar labels are connected, iid data increases cross-client edges and thus communication. Similarly, increasing the number of clients will naturally lead to more crossing edges and proportionally higher overhead (Theorem~\ref{thm:1}), as shown in Figure~\ref{fig:3}.

\section{Conclusion}\label{sec:conclusion}

In this work, we present FedGAT, an algorithm that enables the approximate federated training of Graph Attention Networks for node classification with a single round of pre-training communication. The algorithm can approximate the GAT update equation to arbitrary precision, and we derive bounds on the approximation error. By making use of a polynomial expansion of certain key terms in the GAT update, we generate expressions that remain constant throughout training, and thus, need to be computed and shared only once, reducing communication overhead. These expressions do not leak individual node features, and protect privacy. Our experiments demonstrate that models trained using FedGAT attain nearly the same accuracy as GATs trained in the centralized setting, and that FedGAT is unaffected by the number of clients or data distribution. To the best of our knowledge, FedGAT is one of the first algorithms that computes GAT updates in a near exact manner without dropping cross-client edges, and has rigorous error bounds.


We finally discuss several fascinating research directions and applications of FedGAT that could be explored. 

\textbf{GATs for more complex tasks: }With FedGAT, it is possible to train GATs. This can be very useful for difficult tasks such as fraud detection in payment networks, or predicting the spread of diseases such as COVID-19. Data regulations require payment data to be stored locally, and disease spread prediction requires access to sensitive data, but with FedGAT, it could become possible to perform such complex, large-scale tasks with advanced GAT models.

\textbf{Overcoming data inequity: }Our experiments show that the performance of a model trained with FedGAT is not affected by the distribution of data. This could be useful in a practical setting with naturally non-iid data.




\textbf{Node degree-agnostic feature aggregation: } FedGAT complexity is strongly affected by the maximum node degree. Optimizing the feature aggregation and communication method to be agnostic to the node degree could go a long way in making the algorithm much more scalable.


\textbf{Enabling multi-hop GAT/GNN training: } As currently designed, FedGAT is not optimized to handle GATs with several layers (more than 2). As learning tasks become more complex, and graphs more connected, it has become necessary to find efficient ways to handle multi-hop GNNs.

\textbf{Theoretical privacy guarantees: } Like FedGCN, FedGAT relies on feature aggregation to preserve node privacy across clients. Further formalizing these guarantees and empirically verifying them would strengthen the privacy evaluation of FedGAT.



\nocite*{}

\bibliography{aaai25.bib}

\clearpage
\appendix

\section{FedGAT Algorithm}
We show the pre-training process of FedGAT in Algorithm~\ref{algo:pre-train} and the federated training process in Algoritm~\ref{algo:fedgat-train}.

\begin{algorithm}[ht]
    \caption{FedGAT pre-training round}\label{alg:1}
    \begin{algorithmic}
        \State{Attention parameters $a_{1}, a_{2}$, weight parameter $W$ for each head of each layer of GAT satisfy $\|a_{1}\|, \|a_{2}\|, \|W\|\leq 1$.}
        \State{Server $\mathcal{S}$, clients $\{\mathcal{C}_{k}\}_{k = 1}^{K}$, nodes $i = 1, 2, \cdots, N$.}
        \State{Graph $\mathcal{G} = (\mathcal{V}, E)$; $\vert \mathcal{V}\vert = N$.}
        \State{Node subset at client $\mathcal{C}_{k}$ is $\mathcal{V}_{k}$, and induced sub-graph $\mathcal{G}_{k}$.}
        \State{Feature vectors $h_{i}\in\mathbf{R}^{d}$}
        \State{$\mathcal{S}$ requests all nodes to transmit node features $\left\{ h_{i} \right\}_{i = 1}^{N}$}
        \State{At server $\mathcal{S}$}
        \For{node $i = 1$ to $N$}
        \State{Generate orthonormal vectors $\left\{u_{1j}, u_{2j}\right\}_{j\in\mathcal{N}_{i}}$.}
        \State{$\mathcal{U}_{j} = \frac{1}{2}\left( u_{1j}u_{1j}^{T} + u_{2j}u_{2j}^{T} + \frac{1}{r}u_{1j}u_{2j}^{T} + ru_{2j}u_{1j}^{T}\right)$ for $j\in\mathcal{N}_{i}$ (Eq.\ref{eq:9})}
        \State{$M_{1i}(s) = \sum_{j\in\mathcal{N}_{i}}h_{i}(s)\mathcal{U}_{j}$ for $s = 1, \cdots d$}
        \State{$M_{2i}(s) = \sum_{j\in\mathcal{N}_{i}}h_{j}(s)A_{j}$ for $s = 1, \cdots d$}
        \State{$K_{1i} = \sqrt{2}\sum_{j\in\mathcal{N}_{j}}u_{1j}$, $K_{2i} = \sqrt{2}\sum_{j\in\mathcal{N}_{i}}u_{1j}h_{j}^{T}$}
        \EndFor
        \State{$\mathcal{S}$ transmits $\left\{\{M_{1i}(s), M_{2i}(s)\}_{s = 1}^{d}, K_{1i}, K_{2i}\right\}_{i = 1}^{N}$} to the nodes.
        \State{Node $i$ receives $\{M_{1i}(s), M_{2i}(s)\}_{s = 1}^{d}, K_{1i}, K_{2i}$.}
    \end{algorithmic}
    \label{algo:pre-train}
\end{algorithm}

\begin{algorithm}[ht]
    \caption{FedGAT model training algorithm}\label{alg:2}
    \begin{algorithmic}
        \State{Server $\mathcal{S}$, clients $\mathcal{C}_{k}$ $k = 1, \cdots K$.}
        \State{Server model parameters $\mathcal{W}$, client $\mathcal{C}_{k}$ parameters $\mathcal{W}_{k}$, $T$ global training rounds}
        \State{GAT model - $L$ layers, $\Omega_{l}$ heads at layer $l$; learnable parameters $a_{1}^{(l)}(\omega), a_{2}^{(l)}(\omega), W^{(l)}(\omega)$ at attention head $\omega = 1, \cdots, \Omega(l)$ at layer $l$ of GAT.}
        \For{$t = 1, \cdots, T$ rounds}
        \State{Update parameters at each client with server parameters -$\mathcal{W}_{k}^{(t - 1)}\leftarrow \mathcal{W}^{(t - 1)}$.}
        \State{Select subset of clients $CS(t)$.}
        \State{At each client $\mathcal{C}_{k}\in CS(t)$, compute GAT updates - }
        \For{GAT layer $l = 1, \cdots, L$}
            \If{$l = 1$}
                \State{For all nodes $i\in\mathcal{C}_{k}$, and all attention heads - }
                \State{$b_{1} = W^{T}a_{1}, b_{2} = W^{T}a_{2}$.}
                \State{$D_{i} = \sum_{s = 1}^{d}b_{1}M_{1i}(s) + b_{2}M_{2i}(s)$ (Eq. \ref{eq:14})}
                \State{$E_{i}^{(n)} = \left(K_{1i}^{T}D_{i}^{n}K_{2i}\right)^{T}$ 
 (Eq.\ref{eq:12})}
                \State{$F_{i}^{(n)} = K_{1i}^{T}D_{i}^{n}K_{1i}$ 
 (Eq.\ref{eq:12})}
                \State{Update using series approximation (Eq.\ref{eq:7})}
            \Else
            \State{Regular GAT equations (Eq.\ref{eq:1}, \ref{eq:2}, \ref{eq:3}).}
            \EndIf
        \EndFor
        \State{Update model parameters $\mathcal{W}_{k}^{(t)}\leftarrow \mathcal{W}_{k}^{(t - 1)}$ at each client $\mathcal{C}_{k}\in CS(t)$.}
        \State{Aggregate at server $\mathcal{W}^{(t)} = \frac{1}{\|CS(t)\|}\sum_{k\in CS(t)}\mathcal{W}_{k}^{(t)}$}
        \EndFor
    \end{algorithmic}
    \label{algo:fedgat-train}
\end{algorithm}
\section{Background}

\subsection{Federated Learning}

Federated learning has become a common framework for training machine learning models, in which a set of clients performs local updates by gradient descent using their lo-
cal data and shares the optimization parameters with a co-ordinating server to keep the local client data secure. Several algorithms have been proposed, including, but not limited to \textbf{FedAvg}~\cite{mcmahan2016federated}, \textbf{FedProx}~\cite{li2020federatedoptimizationheterogeneousnetworks}, \textbf{ADMM}~\cite{GABAY197617,Glowinski1975}, \textbf{FedAdam}~\cite{reddi2020adaptive}, \textbf{APFL}~\cite{deng2020adaptive}. 

In this work, we have primarily focused on FedAvg as the algorithm of choice to perform parameter aggregation. At global training round $t$, the server sends a copy of the global model parameters $\mathcal{W}^{(t)}$ to all the clients. Each client then performs $E$ local iterations, at iteration computing a gradient $g^{(t, e)}_{k}$ using the local data to update its local parameters:

\begin{equation}
    \mathcal{W}_{k}^{(t, e + 1)} = \mathcal{W}_{k}^{(t, e)} - \eta g_{k}^{(t, e + 1)}\label{eq:19}
\end{equation}

After all clients complete $E$ local iterations, the global server aggregates the local model parameters and performs an update as

\begin{equation}
    \mathcal{W}^{(t + 1)} = \frac{1}{K}\sum_{k = 1}^{K}\mathcal{W}_{k}^{(t, E)}\label{eq:20}
\end{equation}

The process continues till a stopping criterion is reached.

\subsection{Graph Neural Networks}

Graph Neural Networks are a family of models that learn from graph-structured data. A typical GNN is obtained by stacking multiple GNN layers. The typical GNN update equation for the $l^{th}$ layer is as follows:

\begin{equation}
    h_{i}^{(l)} = \phi\left( \text{aggregate}_{j\in\mathcal{N}_{i}}\left( \psi(h_{j}^{(l - 1)}) \right)\right)
\end{equation}

Some of the popular GNN models are \textbf{GCN}~\cite{kipf2017semisupervisedclassificationgraphconvolutional}, \textbf{GAT}~\cite{veličković2018graphattentionnetworks}, \textbf{GraphSage}~\cite{hamilton2018inductiverepresentationlearninglarge}. In this work, we exclusively focus on Graph Attention Networks (GATs). The GAT update is

$$H^{(l)} = \phi(\alpha^{(l)} H^{(l = 1)}W^{(l)})$$

where $\alpha$ is the matrix of attention coefficients, computed as

$$\alpha_{ij} = \frac{e_{ij}}{\sum_{k\in\mathcal{N}_{i}}e_{ik}}$$

and 

$$e_{ij} = \exp\bigg(\psi(a_{1}^{T}h_{i}^{(l - 1)} + a_{2}^{T}h_{j}^{(l - 1)})\bigg)$$

$H^{(l)}$ is the matrix of embeddings of the nodes after the $l^{th}$ GAT layer. The embeddings of an $L$-layer GAT depend on the $L$-hop neighbours of the nodes.

\subsection{Chebyshev Series}

In function approximation, the Chebysehv Polynomials are an important tool. The Chebyshev polynomials of the first kind are defined as follows:

\begin{equation}
    T_{n}(\cos(\theta)) = \cos(n\theta)\label{eq:22}
\end{equation}

For a function defined on $[-1, 1]$, the set of Chebyshev polynomials form an orthonormal basis (under an appropriate weight function) of continuous functions $f:[-1, 1]\rightarrow\mathbf{R}$.

The Chebyshev polynomials satisfy

\begin{equation}
    \int_{-1}^{1}\frac{T_{n}(x)T_{m}(x)}{\sqrt{1 - x^{2}}}dx = \begin{cases}
        0 & n\neq m\\
        \pi & n = m = 0\\
        \frac{\pi}{2} & n = m \neq 0\\
    \end{cases}\label{eq:23}
\end{equation}

The Chebyshev series of a function $f$ is defined as the series where

$$f(x) = \sum_{n = 0}^{\infty}a_{n}T_{n}(x)$$

and $$a_{n} = \begin{cases}
    \frac{2}{\pi}\int_{-1}^{1}\frac{f(x)T_{n}(x)}{\sqrt{1 - x^{2}}}dx & n > 0\\
    \frac{1}{\pi}\int_{-1}^{1}\frac{f(x)T_{n}(x)}{\sqrt{1 - x^{2}}}dx & n = 0\\
\end{cases}$$

This set of polynomials is widely used because they can be easily computed, and have good error convergence rate in function approximation. It turns out that the truncated Chebyshev series of degree $n$ is a near optimal polynomial approximation in the max-norm. For a function $f:[-1, 1]\rightarrow\mathbf{R}$ that is $k$-times differentiable, the approximation error in the max-norm of the truncated Chebyshev series of degree $n>k$ decays as $\mathcal{O}(\frac{1}{k(n - k)^k})$ \cite{trefethen1981rational, rivlin2020chebyshev}.

\section{Settings and Further Experiments}

In this section, we provide the detailed settings and some more experiments that we have completed to support the algorithm.

\subsection{Computational Device(s) and Settings for Experiments}

\begin{table*}[t]
    \centering
    \begin{tabular}{|M{3cm}|M{2cm}|M{2cm}|M{2cm}|}
    \hline
      \textbf{Attribute} & \textbf{Cora}  &  \textbf{Citeseer} & \textbf{Pubmed}\\
       \hline
       Nodes & 2708 & 3327 & 19717 \\
       \hline
       Edges & 5429 & 4732 & 44338 \\
       \hline
       Feature dim & 1433 & 3703 & 500 \\
       \hline
       Classes & 7 & 6 & 3 \\
       \hline
       Train & 140 & 120 & 60 \\
       \hline
       Validation & 500 & 500 & 500 \\
       \hline
       Test & 1000 & 1000 & 1000 \\
       \hline
    \end{tabular}
    \caption{Summary of datasets used for experiments.}
    \label{tab:2}
\end{table*}

For all experiments, we used an approximation scheme that was a degree 16 approximation.

For Cora, we used a GAT with 2 layers, the number of hidden dimensions was $8$, and the number of attention heads was $8$. The optimiser used was Adam \cite{kingma2017adammethodstochasticoptimization}, with a model regularisation value of $0.001$, and a learning rate of $0.1$.

For Citeseer, we used a GAT with the exact same architecture as Cora.

For Pubmed, the architecture was changed to include 8 attention heads in the output layer too. The rest of the architecture was kept the same.

The experiments were conducted on an AWS m5.16xlarge instance with a total memory of 256 GiB.

\subsection{Further Experiments}

We then discuss some of the other experiments we run.

\subsubsection{Communication Cost Evaluation}

\begin{figure}[t]
\centering
\includegraphics[width=0.6\linewidth]{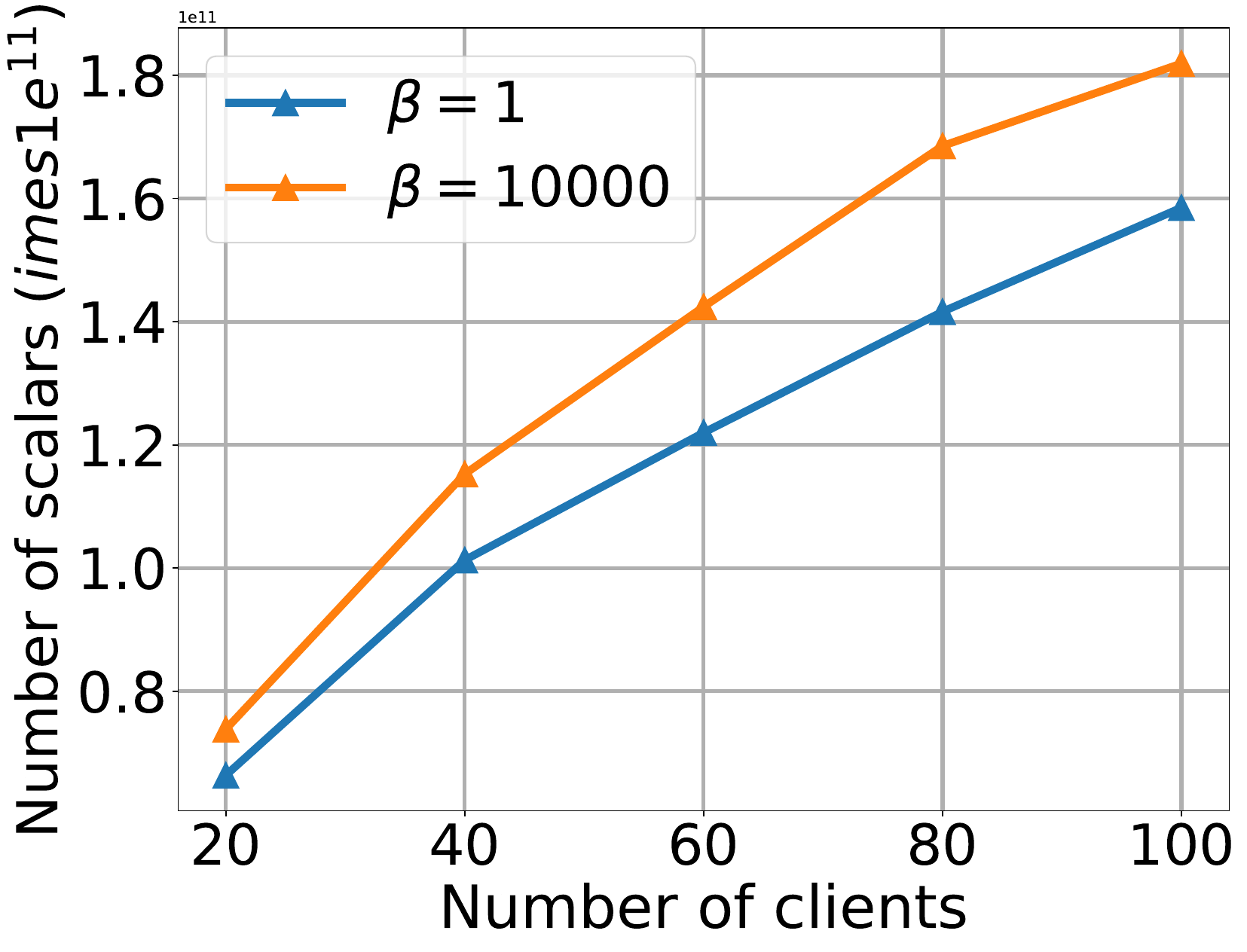}
    \caption{Pretrain communication cost for the \textbf{Cora} dataset. Here, the number of clients ranges from 20 to 100. Once again, we observe a near linear increase in the cost with the number of clients.}
    \label{fig:4}
\end{figure}

To further explore the behaviour of the pretrain communication cost, we computed the cost for the Cora dataset with a large number of clients. We observed nearly the same pattern as with a smaller number of clients. The cost increases with the number of clients. Additionally, the iid setting ($\beta = 10000$) resulted in more communication cost than the non-iid setting ($\beta = 1$). This further strengthens our intuition that iid distribution of data leads to more cross-client edges, thus resulting in larger sub-graphs at each client. The results of the computation are shown in Figure \eqref{fig:4}.

\subsubsection{Variation in Accuracy with Approximation Degree}

In this section, we briefly look at the variation in the test accuracy of FedGAT when the degree of approximation for the Chebyshev series of attention scores is varied. 

\begin{figure*}[t]
    \centering
    \begin{subfigure}[b]{0.23\linewidth}
    \centering
        \includegraphics[width = \linewidth]{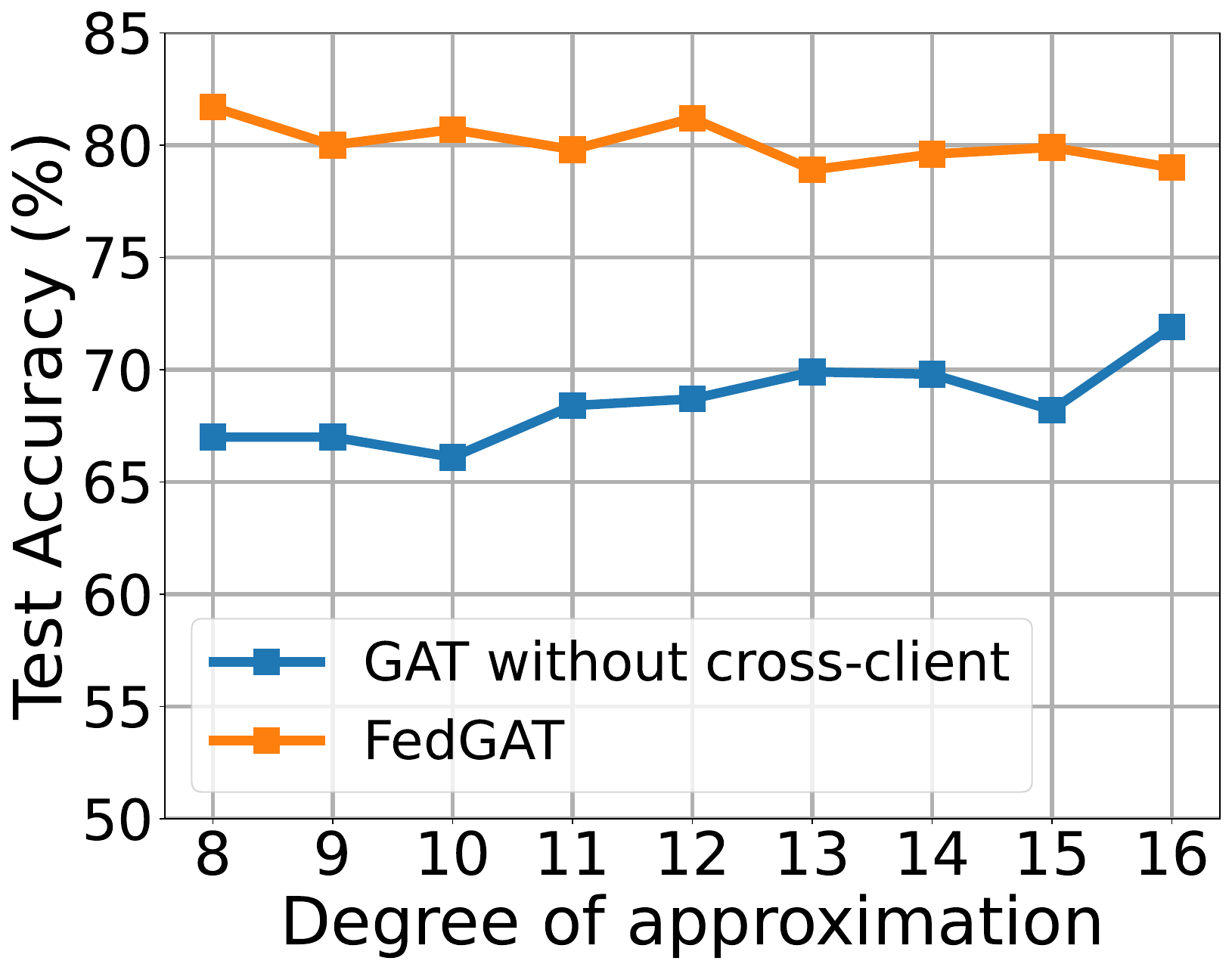}
        \caption{$\beta = 1$ (non-iid)}
    \end{subfigure}
    \centering
    \begin{subfigure}[b]{0.23\linewidth}
    \centering
        \includegraphics[width = \linewidth]{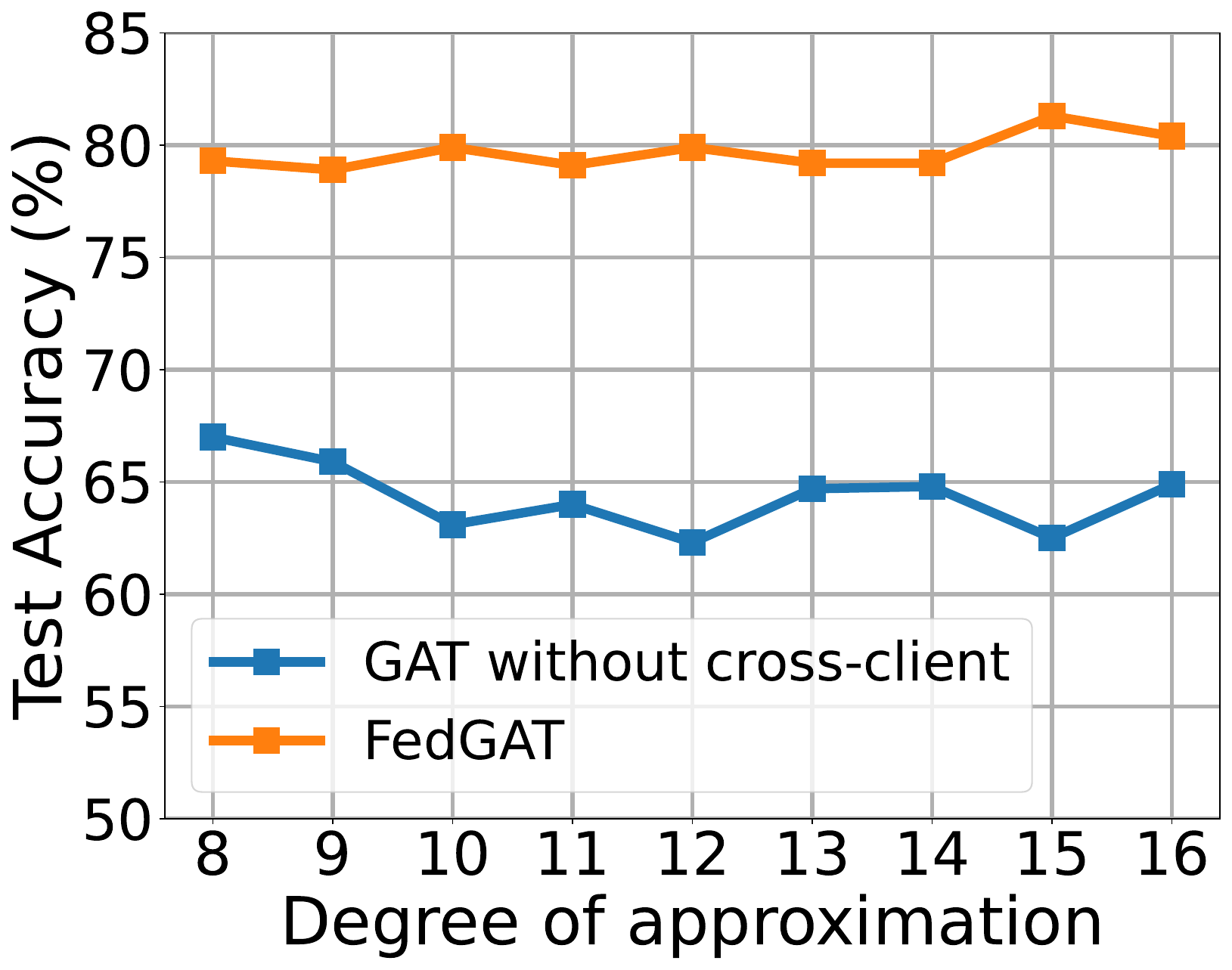}
        \caption{$\beta = 100$ (partial iid)}
    \end{subfigure}
    \centering
    \begin{subfigure}[b]{0.23\linewidth}    
        \includegraphics[width = \linewidth]{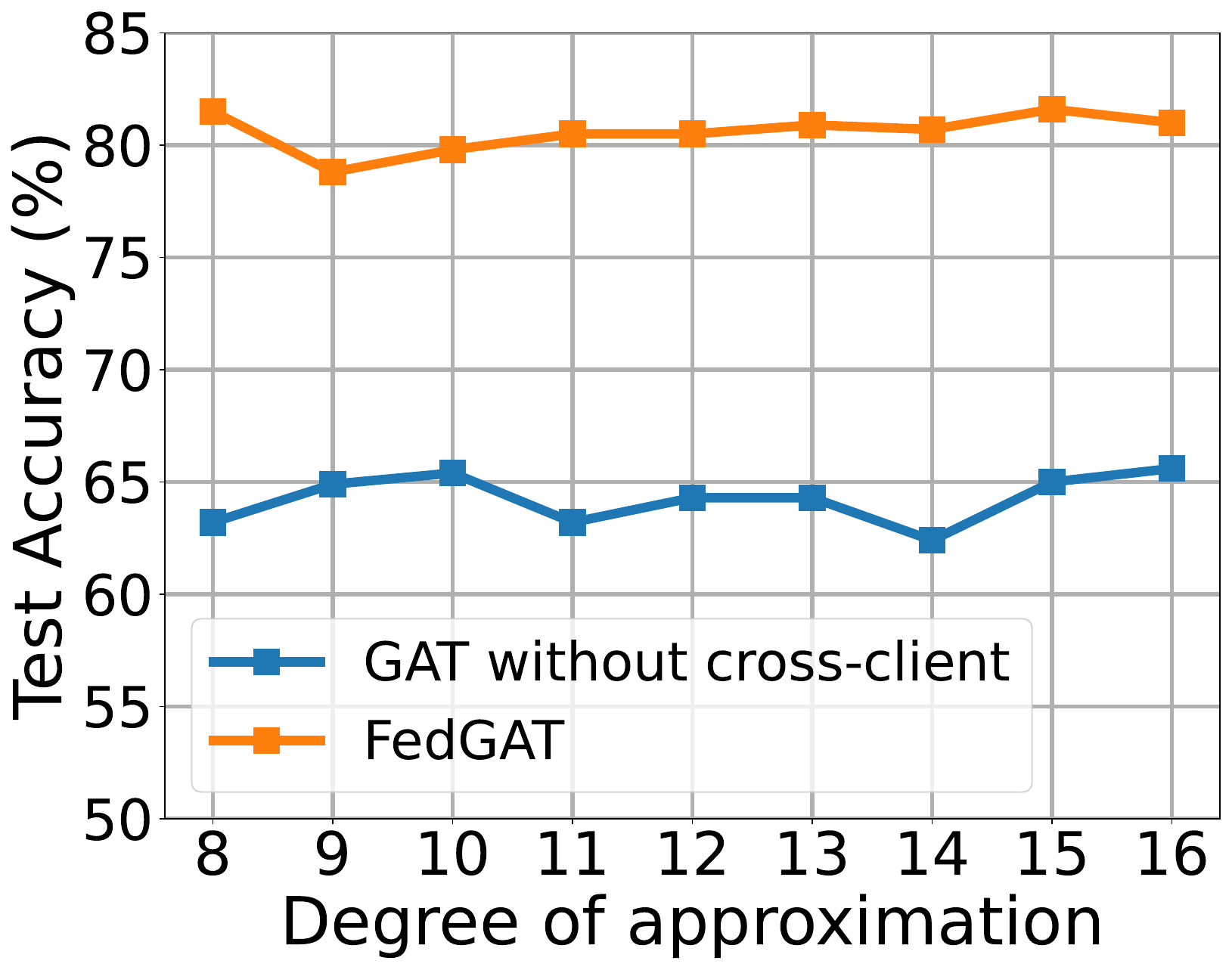}
        \caption{$\beta = 10000$ (iid)}
    \end{subfigure}
    \caption{Accuracy v/s degree of approximation for iid, partial iid and non-iid data distribution on \textbf{Cora}. The performance of DistGAT is shown only for reference; it does not involve approximations.}
    \label{fig:5}
\end{figure*}

We observed (Figure \eqref{fig:5}) that the performance did not change significantly. We attribute this to the fact that the Chebyshev error drops quite fast; therefore, the FedGAT update is very close to the true GAT update even for low degree of approximation.

It is still quite surprising to see the model perform well even for degree $8$; we do not know exactly why this happens, but we suspect that the model has some noise in it due to the low approximation accuracy of the Chebyshev series, leading to better generalisation.

\section{Communication Cost of FedGAT}

Here, we derive the complexity bound on the communication cost of the FedGAT algorithm. 

Let the graph contain $N$ nodes. $B$ denotes the maximum degree of the nodes in the graph. The node features have dimension $d$, and there are $K$ clients. The $L$-hop neighbourhood of a node is the set of all nodes that are at a distance of $L$ from the node. We define the $L$-hop neighbourhood of a sub-graph as the union of all the $L$-hop neighbourhoods of the nodes of the sub-graph. Let $B_{L}$ denote the size of the largest of the $L$-hop neighbourhoods of the client graphs. Finally, assume that a degree $p$ approximation is used to compute the GAT update. Here, we prove Theorem \eqref{thm:1}.

\begin{claim}
The communication cost of the FedGAT algorithm scales as $\mathcal{O}(KB_{L}dB^{2})$. (Theorem \eqref{thm:1})

\end{claim}

\textbf{Proof: }

In the FedGAT algorithm, the server computes the set of matrices $\{M_{1i}(r), M_{2i}(r)\}_{r = 1}^{d}$ for all nodes $i$ in the graph. The matrices are constructed as

$$M_{1i}(r) = h_{i}(r)\sum_{j\in\mathcal{N}_{j}}\mathcal{U}_{j}$$

where $\mathcal{U}_{j}$ is defined in Eq. \eqref{eq:9}. Note that the way $\mathcal{U}_{j}$ are defined for a given node $i$, $\mathcal{U}_{j}$ must have dimensions at least $2deg(i)\times 2deg(i)$, where $deg(i)$ denotes the degree of node $i$. This is because we construct a set of orthonormal vectors with $2deg(i)$ vectors in it. This can be achieved by using vectors of size $2deg(i)$. Since matrices $\mathcal{U}_{j}$ are a result of the outer product between vectors from this set, they must have dimensions $2deg(i)\times2deg(i)$. 

Now, the rest of the proof is clear. The server computes $d$ matrices of the aforementioned dimensions for node $i$. Thus, the number of scalars to be communicated to node $i$ is $4deg(i)^{2}\times d$. Now, for client $\mathcal{C}_{k}$, the size of the $L$-hop subgraph is upper bounded by $B_{L}$. Thus, there are at most $B_{L}$ nodes at the client. The total communication cost for this client is $\sum_{i\in\\mathcal{C}_{k}}4deg(i)^{2}\times d = \mathcal{O}(B_{L}dB^{2})$. Sine there are $K$ clients, the total communication cost is $\mathcal{O}(KB_{L}dB^{2})$.

It is pertinent to note that for social network graphs, the diameter is often of the order of $O(\log N)$, where $N$ is the number of nodes in the graph. It is often possible that for $L$ sufficiently large, the $L$-hop neighbourhood is of size $\mathcal{O}(N)$. In such a case, $B_{L} = \mathcal{O}(N)$, and the FedGAT algorithm will scale as $\mathcal{O}(KNdB^{2})$.

\section{Proofs of Approximation Error}

In this section, we prove Theorems \eqref{thm:3}, \eqref{thm:4} and \eqref{thm:5} that establish results about the approximation error in the FedGAT algorithm. As before, we shall adopt the following notation.

Let the true attention coefficients be $\alpha_{ij}$, and the approximate coefficients be $\hat{\alpha}_{ij}$. Similarly, let $e_{ij}$, as defined in \eqref{eq:3}, be the true attention scores, and $\hat{e}_{ij}$ be the approximate score. Note that the Chebyshev series is used for the computation of the scores $\hat{e}_{ij}$, which are then used in computing $\alpha_{ij}$.

\begin{claim}
    In the single layer FedGAT, let $\|e_{ij} - \hat{e}_{ij}\|_{2} \leq \epsilon$ $\forall i, j$. Then, $\|\alpha_{ij} - \hat{\alpha}_{ij}\|_{2} \leq \alpha_{ij}\frac{2\epsilon}{1 - \epsilon}$.
\end{claim}

\textbf{Proof: }

Recall that $\alpha_{ij} = \frac{e_{ij}}{\sum_{k\in\mathcal{N}_{i}}e_{ik}}$. Then, $\hat{\alpha}_{ij} = \frac{\hat{e}_{ij}}{\sum_{k\in\mathcal{N}_{i}}\hat{e}_{ik}}$. Since $\|e_{ij} - \hat{e}_{ij}\|_{2} \leq \epsilon$, we can re-write this as $\hat{e}_{ij} = \Delta_{ij}e_{ij}$, so that $\vert e_{ij}\vert\|\Delta_{ij} - 1\|\leq \epsilon$. It is clear that $e_{ij}$ is always a positive number. Then, we can write 

$$\hat{\alpha}_{ij} = \frac{e_{ij}\Delta_{ij}}{\sum_{k\in\mathcal{N}_{i}}e_{ik}\Delta_{ik}}$$

$$\|\alpha_{ij} - \hat{\alpha}_{ij}\| = \left\|\frac{e_{ij}}{\sum_{k\in\mathcal{N}_{i}}e_{ik}} - \frac{e_{ij}\Delta_{ij}}{\sum_{k\in\mathcal{N}_{i}}e_{ik}\Delta_{ik}}\right\|$$

$$
\begin{aligned}
    \|\alpha_{ij} - \hat{\alpha}_{ij}\| &= \vert\frac{e_{ij}}{\sum_{k\in\mathcal{N}_{i}}e_{ik}}\vert\left\|\frac{\sum_{k\in\mathcal{N}_{i}}e_{ik}(\Delta_{ik} - \Delta_{ij})}{\sum_{k\in\mathcal{N}_{i}}e_{ik}\Delta_{ik}}\right\|\\
    &= \alpha_{ij}\left\|\frac{\sum_{k\in\mathcal{N}_{i}}e_{ik}\Delta_{ik}(1 - \frac{\Delta_{ij}}{\Delta_{ik}})}{\sum_{k\in\mathcal{N}_{i}}e_{ik}\Delta_{ik}}\right\|\\
    &= \alpha_{ij}\left\|\sum_{k\in\mathcal{N}_{i}}\hat{\alpha}_{ij}(1 - \frac{\Delta_{ij}}{\Delta_{ik}})\right\|\\
    &\leq \alpha_{ij}\max_{i, j, k}\left\|1 - \frac{\Delta_{ij}}{\Delta_{ik}}\right\|
\end{aligned}$$

Now, we need to bound the maximum of $\|1 - \frac{\Delta_{ij}}{\Delta_{ik}}\|$. Since $e_{ij}\|\Delta_{ij} - 1\| \leq \epsilon$, and $e_{ij} = \exp\left(\psi(a_{1}^{T}h_{i} + a_{2}^{T}h_{j})\right)$, with $a_{1}, a_{2}, h_{i}, h_{j}$ all being norm-bounded (Assumptions \eqref{assm:2}, \eqref{assm:3}), we know that $e_{ij}\geq 1$. Therefore, $\|\delta_{ij} - 1\| \leq \epsilon$.

Therefore,

$$1 - \epsilon \leq \|\Delta_{ij}\| \leq 1 + \epsilon$$

Thus, $\max\left\|1 - \frac{\Delta_{ij}}{\Delta_{ik}}\right\| \leq \max\left\{\frac{1 + \epsilon}{1 - \epsilon} - 1, 1 - \frac{1 - \epsilon}{1 + \epsilon}\right\| = \frac{2\epsilon}{1 - \epsilon}$. Therefore,

$$\|\alpha_{ij} - \hat{\alpha}_{ij}\| \leq \alpha_{ij}\frac{2\epsilon}{1 - \epsilon}$$

Hence proved.

\begin{claim}
    Let $h_{i}^{(1)}$ be the embedding of node $i$ obtained after the first GAT layer, and $\hat{h}_{i}^{(1)}$ be the embedding generated by the FedGAT layer.

    If $\|e_{ij} - \hat{e}_{ij}\|\leq \epsilon$, then,

    $$\|h_{i}^{(1)} - \hat{h}_{i}^{(1)}\| \leq \frac{2\kappa_{\phi}\epsilon}{1 - \epsilon}$$
\end{claim} 

\textbf{Proof: }

Recall the GAT update equation as given in Eq. \eqref{eq:1}.

$$h_{i}^{(1)} = \phi\left(\sum_{j\in\mathcal{N}_{i}}\alpha_{ij}Wh_{j}^{(0)}\right)$$

Re-expressing the attention coefficients $\alpha_{ij}$ in terms of the scores $e_{ij}$ yields the GAT update as

$$h_{i}^{(1)} = \phi\left(\frac{\sum_{j\in\mathcal{N}_{i}}e_{ij}Wh_{j}^{(0)}}{\sum_{k\in\mathcal{N}_{i}}e_{ik}}\right)$$

Since the FedGAT algorithm approximates the attention scores $e_{ij}$ using a Chebyshev series, the update remains the same, but the scores are now approximate.

Thus,

$$\hat{h}_{i}^{(1)} = \phi\left(\frac{\sum_{j\in\mathcal{N}_{i}}\hat{e}_{ij}Wh_{j}^{(0)}}{\sum_{k\in\mathcal{N}_{i}}\hat{e}_{ik}}\right)$$

Therefore,

\[\begin{aligned}
    \|h_{i}^{(1)} - \hat{h}_{i}^{(1)}\| &=
\bigg\|\phi\left(\frac{\sum_{j\in\mathcal{N}_{i}}\hat{e}_{ij}Wh_{j}^{(0)}}{\sum_{k\in\mathcal{N}_{i}}\hat{e}_{ik}}\right)\\ 
&- \phi\left(\frac{\sum_{j\in\mathcal{N}_{i}}e_{ij}Wh_{j}^{(0)}}{\sum_{k\in\mathcal{N}_{i}}e_{ik}}\right)\bigg\|\\
\end{aligned}\]

All activations $\phi$, $\psi$ used in the algorithm are Lipschitz continuous and monotone. As a consequence, $\forall x, y$, $\|\phi(x) - \phi(y)\| \leq \kappa_{phi}\|x - y\|$. Thus, using this fact and the triangle inequality of norms, we get that

\[\begin{aligned}
    \|h_{i}^{(1)} - \hat{h}_{i}^{(1)}\| &\leq
    \kappa_{\phi}\left\|\sum_{j\in\mathcal{N}_{i}}(\alpha_{ij} - \hat{\alpha}_{ij})Wh_{j}^{(0)}\right\|\\
    &\leq \kappa_{\phi}\sum_{j\in\mathcal{N}_{i}}\|(\hat{\alpha}_{ij} - \alpha_{ij})Wh_{j}\|\\
\end{aligned}\]

We make use of the fact that the norm of the parameters and the features is bounded by $1$. Additionally, we can use the fact that for a matrix $A$ and a vector $x$, $\|Ax\|\leq \|A\|\|x\|$ when using the $L_{2}$ norm for the matrix and the vector.

Thus, 

\[\begin{aligned}
    \|h_{i}^{(1)} - \hat{h}_{i}^{(1)}\| &\leq \sum_{j\in\mathcal{N}_{i}}\|(\hat{\alpha}_{ij} - \alpha_{ij})Wh_{j}\|\\
    &\leq \kappa_{\phi}\sum_{j\in\mathcal{N}_{i}}\|\hat{\alpha}_{ij} - \alpha_{ij}\|\|W\|\|h_{j}^{(0)}\|\\
    &\leq \kappa_{\phi}\sum_{j\in\mathcal{N}_{i}}\alpha_{ij}\frac{2\epsilon}{1 - \epsilon}\\
    &=\frac{2\kappa_{\phi}\epsilon}{1 - \epsilon}\\
\end{aligned}\]

Hence, proved.

Finally, we provide a proof for the final error claim of our work. Theorem \eqref{thm:5} by combining the results of Theorems \eqref{thm:3} and \eqref{thm:4}. This theorem establishes the behaviour of the error as we stack more GAT layers.

Before that however, we shall prove a lemma that will be useful in Theorem \eqref{thm:5}.

\begin{lemm}
    \emph{If $x \leq \log(c)$,  for some $c > 1$, then, }

    $$\exp(x) - 1 \leq cx$$
\end{lemm}

\textbf{Proof: }

We make use of the power series of the function for this purpose.

$$\exp(x) = \sum_{n = 0}^{\infty}\frac{x^{n}}{n!}$$

Now, consider the function

$$f(x) = cx - (\exp(x) - 1)$$

Then, $$f(x) = (c - 1)x - \sum_{n = 1}^{\infty}\frac{x^{n}}{n!}$$

$$\begin{aligned}
    f^{'}(x) &= (c - 1) - \sum_{n = 1}^{\infty}\frac{nx^{n - 1}}{n!}\\
    &= (c - 1) - \sum_{n = 0}^{\infty}\frac{x^{n}}{n!}\\
    &= (c - 1) - (\exp(x) - 1)\\
    &= c - \exp(x)\\
\end{aligned}$$

Notice that $f^{'}(x) > 0$ $\forall x\leq \log(c)$. Now, $f(0) = 0$. Therefore, in the region $0 \geq x\leq \log(c)$, $f$ is a strictly increasing function, and since the value at $0$ is 0, $f$ must be strictly positive for $x\in[0, \log(c)]$ (in fact, it is positive for $x$ larger than $\log(c)$ upto some point). Hence, if $x\leq \log(c)$, $cx > \exp(x) - 1$.

Hence, proved.

\begin{claim}

Let $h_{i}^{(l)}$ be the embeddings of node $i$ obtained after the $l^{th}$ GAT layer. Let $\hat{h}_{i}^{(l)}$ be the embedding obtained after the $l^{th}$ layer in the FedGAT algorithm.

Let $\|h_{i}^{(l - 1)} - \hat{h}_{i}^{(l - 1)}\|\leq \delta$, and let $\delta\leq \frac{\log(c)}{2\kappa_{\psi}}$, where $c$ is a positive constant. Then, the following results hold:

    $$\epsilon  = \|\hat{e}_{ij}^{(l)} - e_{ij}^{(l)}\| \leq 2c\kappa_{\psi}\delta$$

    $$\|\hat{\alpha}_{ij}^{(l)} - \alpha_{ij}^{(l)}\| \leq \alpha_{ij}^{(l)}\frac{2\epsilon}{1 - \epsilon} = \alpha_{ij}^{(l)}\varepsilon$$

    $$\|\hat{h}_{i}^{(l)} - h_{i}^{(l)}\|\leq \kappa_{\phi}(\varepsilon + \delta)$$\label{lem:1}
\end{claim}

\textbf{Proof: }

We first establish the error in the attention coefficients.

Recall that only the first layer is based on the FedGAT algorithm. All layers after that use the regular GAT update. Thus, for $l>1$, the attention score is computed as

$$e_{ij}^{(l)} = \exp\left(\psi\left(a_{1}^{T}h_{j}^{(l - 1)} +a_{2}^{T}h_{j}^{(l - 1)}\right)\right)$$

Therefore, 

$$\hat{e}_{ij}^{(l)} = \exp\left(\psi\left(a_{1}^{T}\hat{h}_{i}^{(l - 1)} + a_{2}^{T}\hat{h}_{j}^{(l - 1)}\right)\right)$$

\[\begin{aligned}
    \|e_{ij}^{(l)} - \hat{e}_{ij}^{(l)}\| &= \bigg\|\exp\left(\psi\left(a_{1}^{T}h_{i}^{(l - 1)} + a_{2}^{T}h_{j}^{(l - 1)}\right)\right) - \\
    &\exp\left(\psi\left(a_{1}^{T}\hat{h}_{i}^{(l - 1)} + a_{2}^{T}\hat{h}_{j}^{(l - 1)}\right)\right)\bigg\|\\
\end{aligned}\]

Let $x_{ij} = a_{1}^{T}h_{i}^{(l)} + a_{2}^{T}h_{j}^{(l)}$. Then, $a_{1}^{T}\hat{h}_{i}^{(l)} + a_{2}^{T}\hat{h}_{j}^{(l)} = x_{ij} + \tau_{ij}$ for some $\tau_{ij}$. Then

\[\begin{aligned}
    \|e_{ij}^{(l)} - \hat{e}_{ij}^{(l)}\| &=
    \left\|\exp\left(\psi(x_{ij} + \tau_{ij})\right) - \exp\left(\psi(x_{ij})\right)\right\|\\
    &= \exp(\psi(x_{ij}))\|\exp\bigg(\psi(x_{ij} + \tau_{ij}) - \psi(x_{ij})\bigg) - 1\|\\
    \end{aligned}\]

Now, since the function $\psi$ is a monotonic Lipschitz continuous function, we can upper bound the term $\exp\bigg(\psi(x_{ij} + \tau_{ij}) - \psi(x_{ij})\bigg)$ by $\exp(\kappa_{\psi}\|\tau_{ij}\|)$.

We first place bounds on the term $\tau_{ij}$.

$$\begin{aligned}
    \|\tau_{ij}\| &= \|a_{1}^{T}(\hat{h}_{i}^{(l - 1)} - h_{i}^{(l - 1)}) + a_{2}^{T}(\hat{h}_{j}^{(l - 1)} - h_{j}^{(l - 1)})\|\\
    &= \|a_{1}^{T}\delta_{i} 
+ a_{2}^{T}\delta_{j}\|\\
\end{aligned}$$

where $\delta_{i} = \hat{h}_{i}^{(l - 1)}$.

Then, 

$$\begin{aligned}
    \|\tau_{ij}\| &\leq \|a_{1}^{T}\delta_{i}\| + \|a_{2}^{T}\delta_{j}\|\\
    &\leq \|\delta_{i}\| + \|\delta_{j}\|\\
    &\leq 2\delta\\
\end{aligned}$$

Here, we made use of the fact that the norm of $a_{1}, a_{2}$ is at most $1$. Therefore, $\|\tau_{ij}\|\leq 2\delta$.

We also assumed that $\delta \leq \frac{\log(c)}{2\kappa_{\psi}}$ for some $c>0$. Thus, making use of Lemma \eqref{lem:1}, we can say that

$$\begin{aligned}
    \exp\left(\kappa_{\psi}\|\tau_{ij}\|\right) &\leq \exp\left(2\kappa_{\psi}\delta\right)\\
    &\leq 2c\kappa_{\psi}\delta\\
\end{aligned}$$

Thus, we have shown that

$$\|e_{ij}^{(l)} - \hat{e}_{ij}^{(l)}\| \leq 2c\kappa_{\psi}\delta$$

For the next part about the attention coefficients, we shall assume that

$$2c\kappa_{\psi}\delta = \epsilon$$

Then, making use of the proof of Theorem \eqref{thm:2}, we can say that

$$\|\alpha_{ij} - \hat{\alpha}_{ij}\|\leq \alpha_{ij}\frac{2\epsilon}{1 - \epsilon}$$

Finally, we turn our attention to the node embedding error.

Consider

$\|h_{i}^{(l)} - \hat{h}_{i}^{(l)}\|$; for all layers beyond $l = 1$, the GAT update equation is used.

Thus,

$$\|h_{i}^{(l)} - \hat{h}_{i}^{(l)}\| = \left\|\phi(\sum_{j\in\mathcal{N_{i}}}\hat{\alpha}_{ij}W^{(l)}\hat{h}_{ij}^{(l - 1)}) - \phi\alpha_{ij}W^{(l)}h_{i}^{(l - 1)})\right\|$$

Since $\phi$ is Lipschitz continuous with parameter $\kappa_{\phi}$, we can upper bound the error by

$$\|h_{i}^{(l)} - \hat{h}_{i}^{(l)}\| \leq \kappa_{\phi}\left\|\sum_{j\in\mathcal{N_{i}}}\hat{\alpha}_{ij}W^{(l)}\hat{h}_{ij}^{(l - 1)} - \alpha_{ij}W^{(l)}h_{i}^{(l - 1)}\right\|$$

Let $\delta_{i} = \hat{h}_{i}^{(l - 1)} - h_{i}^{(l - 1)}$. We also recall that the norm of all model paramters in the FedGAT algorithm is at most $1$. Then,

$$\begin{aligned}
    \|h_{i}^{(l)} - \hat{h}_{u}^{(l)}\| &\leq \kappa_{\phi}\bigg\|W^{(l)}\sum_{j\in\mathcal{N}_{i}}(\hat{\alpha}_{ij} - \alpha_{ij})h_{j}^{(l - 1)} + \hat{\alpha}_{ij}\delta_{j}\bigg\|\\
    &\leq \kappa_{\phi}\|W^{(l)}\|\bigg\|\sum_{j\in\mathcal{N}_{i}}(\hat{\alpha}_{ij} - \alpha_{ij})h_{j}^{(l - 1)}\bigg\|\\
    &+\kappa_{\phi}\|W^{(l)}\|\bigg\|\sum_{j\in\mathcal{N}_{i}}\hat{\alpha}_{ij}\delta_{j}\bigg\|\\
    &\leq \kappa_{\phi}\sum_{j\in\mathcal{N}_{i}}\|(\hat{\alpha}_{ij} - \alpha_{ij})h_{j}^{(l - 1)}\| + \kappa_{\phi}\max_{j}\|\delta_{j}\|\\
    &\leq \kappa_{\phi}\bigg(\delta + \sum_{j\in\mathcal{N}_{i}}\alpha_{ij}\varepsilon\|h_{j}^{(l - 1)}\|\bigg)\\
    &\leq \kappa_{\phi}(\delta + \varepsilon)\\
\end{aligned}$$

Thus, we have shown that

$$\|\hat{h}_{i}^{(l)} - h_{i}^{(l)}\|\leq \kappa_{\phi}(\delta + \varepsilon)$$

Hence, proved.

The above results establish that the error in node features scales as $\kappa_{\phi}(\delta + \varepsilon) = \kappa_{\phi}(\delta + \frac{2\epsilon}{1 - \epsilon})$. Expanding this, we get that the error is of the form $\kappa_{\phi}(\delta + \frac{4c\kappa_{\psi}\delta}{1 - 2c\kappa_{\psi}\delta})$. We made an implicit assumption about the fact that the error $\delta$ is small enough that $2c\kappa_{\psi}\delta$ is still small. For the moment, if we assume this to be true, we can see that $\kappa_{\phi}(\delta + \varepsilon) \approx \kappa_{\phi}(\delta(1  + 4c\kappa_{\psi}))$. If we have an $L$-layer GAT, then, if the FedGAT creates an error of $\delta$ in the first layer, the error at the output will be $\kappa_{\phi}^{L}(1 + 4c\kappa_{\psi})^{L}\delta$. Thus, to ensure that error at the output is less than some $\zeta$, the error in the node embeddings at the first layer must be less than $\frac{\zeta}{\kappa_{\phi}^{L}(1 + 4c\kappa_{\psi})^{L}}$. 

In typical applications, the number of layers $L$ is usually 3 or 4. The activations used are ReLU or ELU, both of which have a Lipschitz parameter of $1$. Thus, the error scaling factor is not so large anymore, and can be kept under check quite easily by keeping the Chebyshev error in attention coefficients low. Since the Theorem \eqref{thm:2} establishes good convergence rates for Chebyshev series, it should be possible to do this without much difficulty. As such, the FedGAT approximation scheme is sound.

\section{Speeding up FedGAT}

In this section, we present an efficient version of the FedGAT algorithm that can be used conditionally. This is because there is a chance of leaking node feature vectors in this method. The privacy of the method depends on the node feature vectors; if they are one-hot encoded, or are somehow known to have some special feature, this method may reveal information. However, if that is not the case, then it is possible to speed up the algorithm by reducing the communication overhead bound established in Theorem \eqref{thm:1}. We dub this version of FedGAT as the \textbf{Vector FedGAT}. The original algorithm is henceforth referred to as the \textbf{Matrix FedGAT}. This is because of the fact that Matrix FedGAT communicates matrices, whereas Vector FedGAT communicates $1$-dimensional vectors.

\subsection{Vector FedGAT}

\begin{figure*}[t]
    \centering
    \begin{subfigure}[b]{0.22\linewidth}
    \centering
        \includegraphics[width = \linewidth]{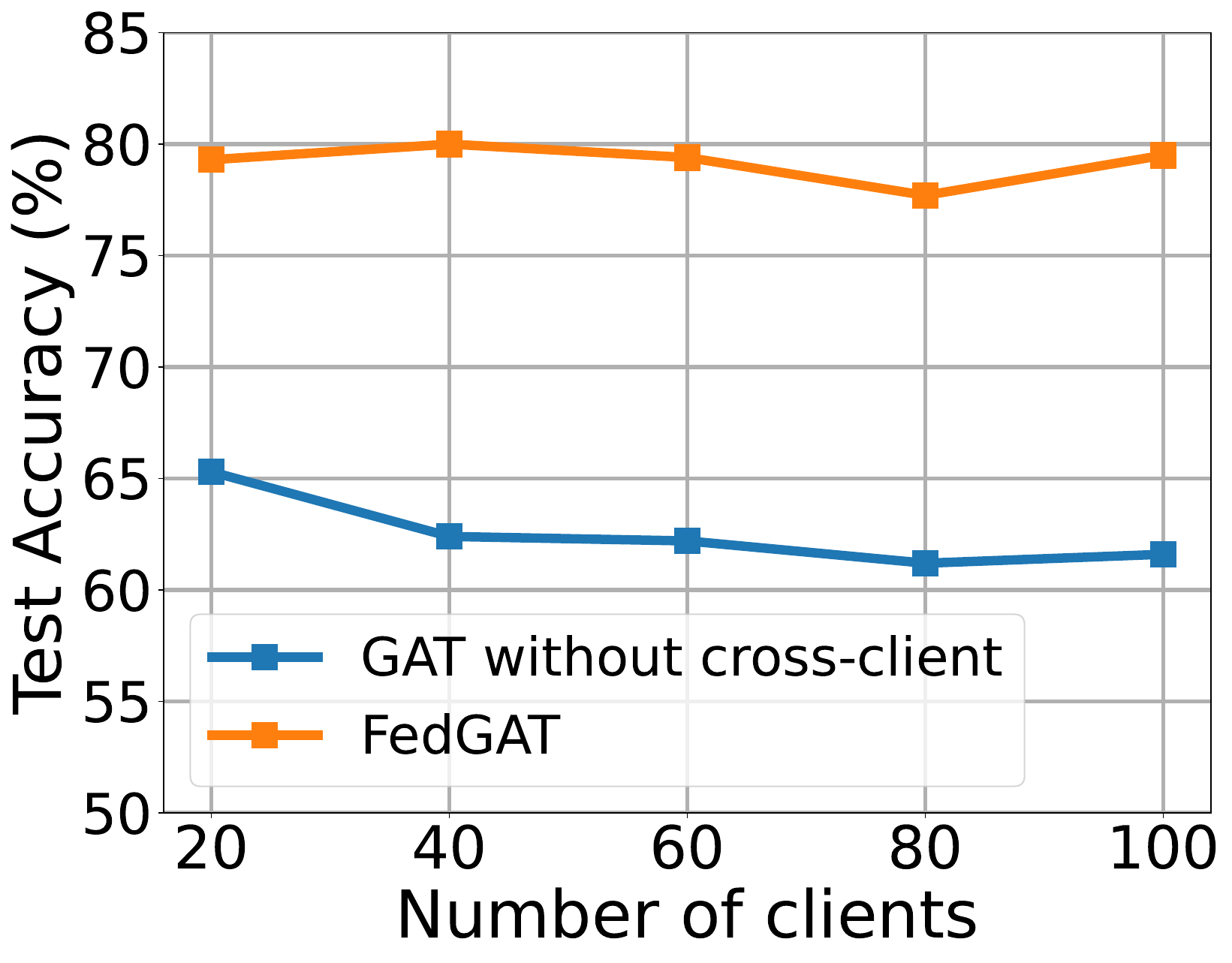}
        \caption{$\beta = 1$ (non-iid)}
    \end{subfigure}
    \centering
    \begin{subfigure}[b]{0.22\linewidth}
    \centering
        \includegraphics[width = \linewidth]{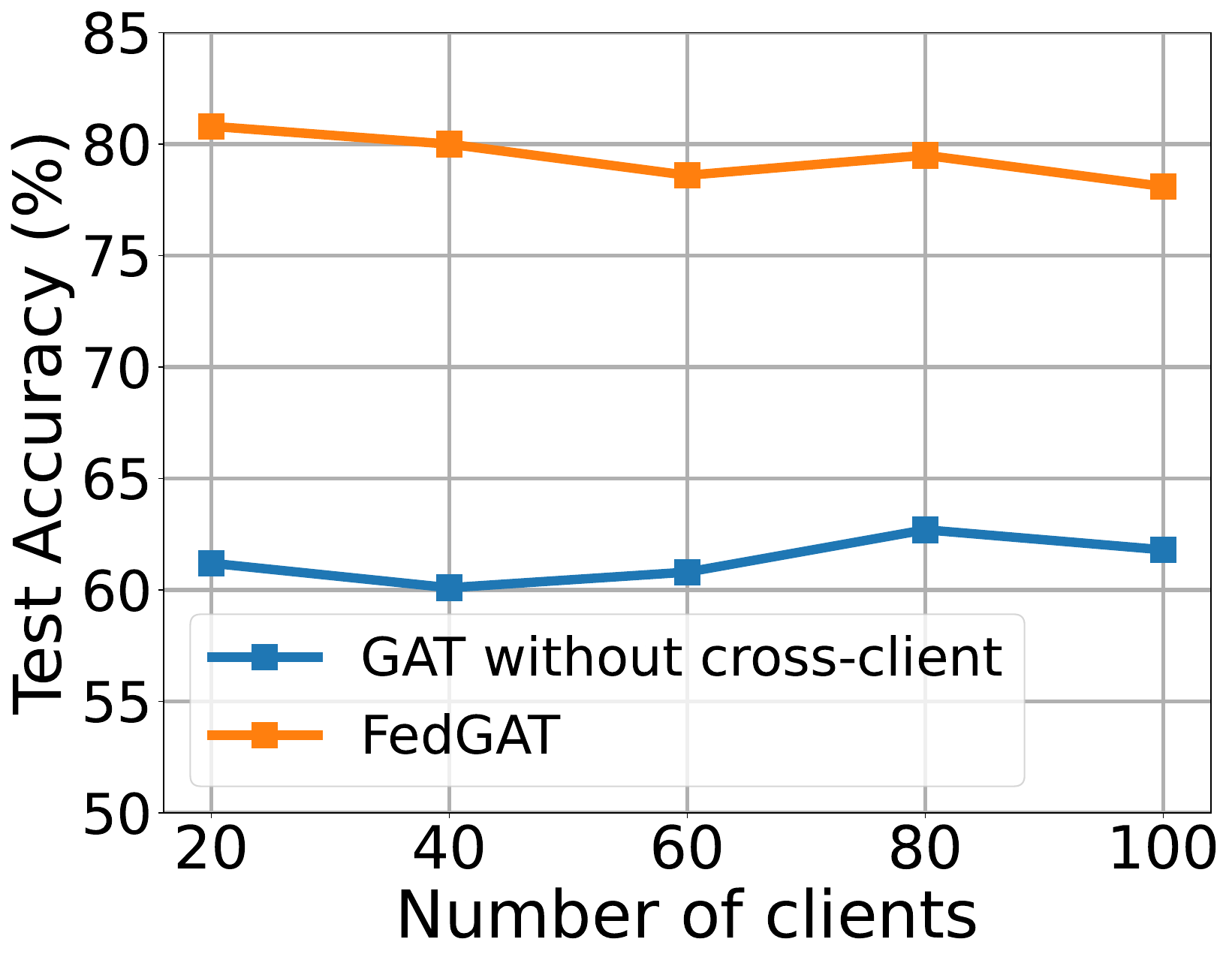}
        \caption{$\beta = 10000$ (iid)}
    \end{subfigure}
    \centering
    \begin{subfigure}[b]{0.22\linewidth}    
        \includegraphics[width = \linewidth]{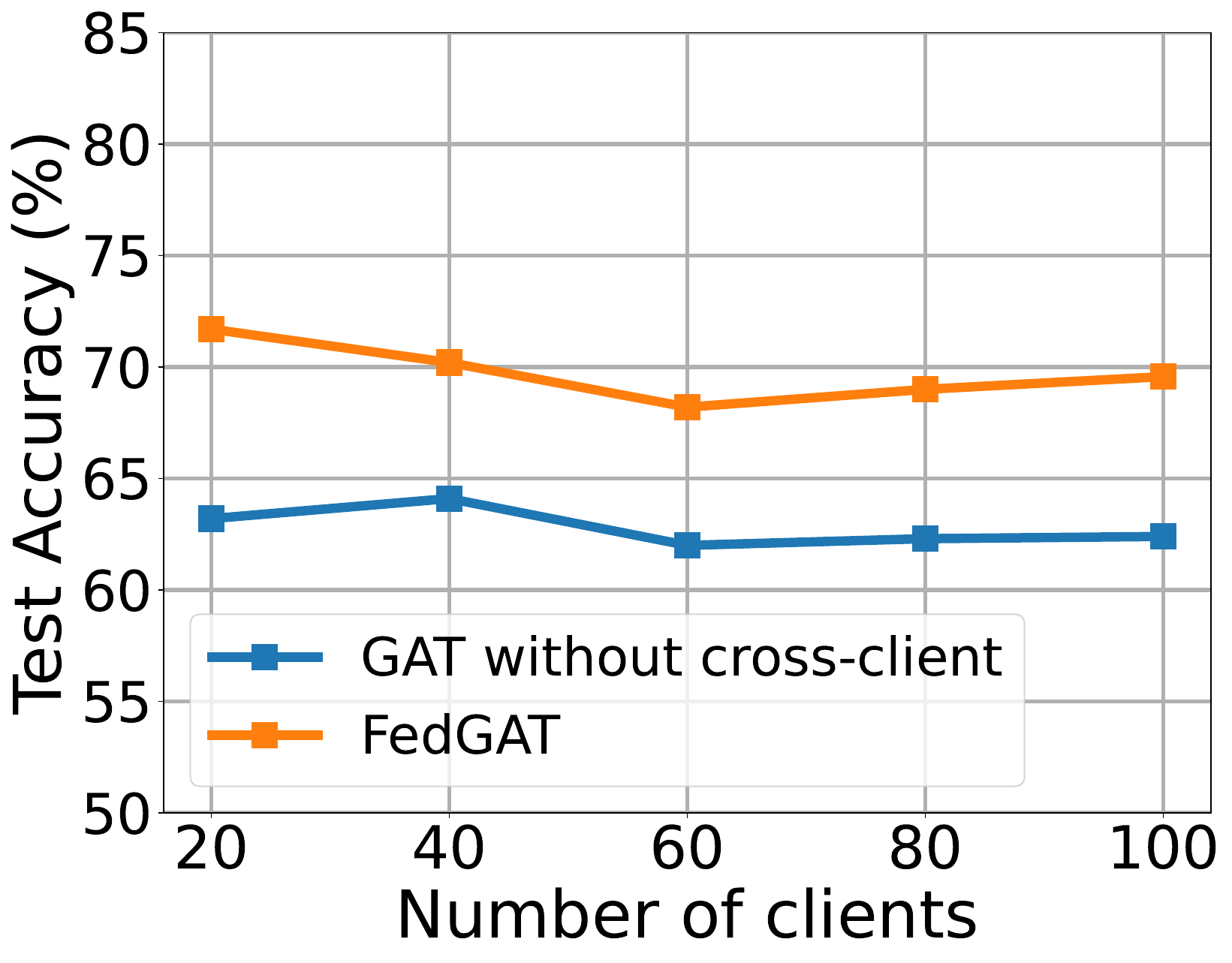}
        \caption{$\beta = 1$ (non-iid)}
    \end{subfigure}
    \centering
    \begin{subfigure}[b]{0.22\linewidth}
    \centering
        \includegraphics[width = \linewidth]{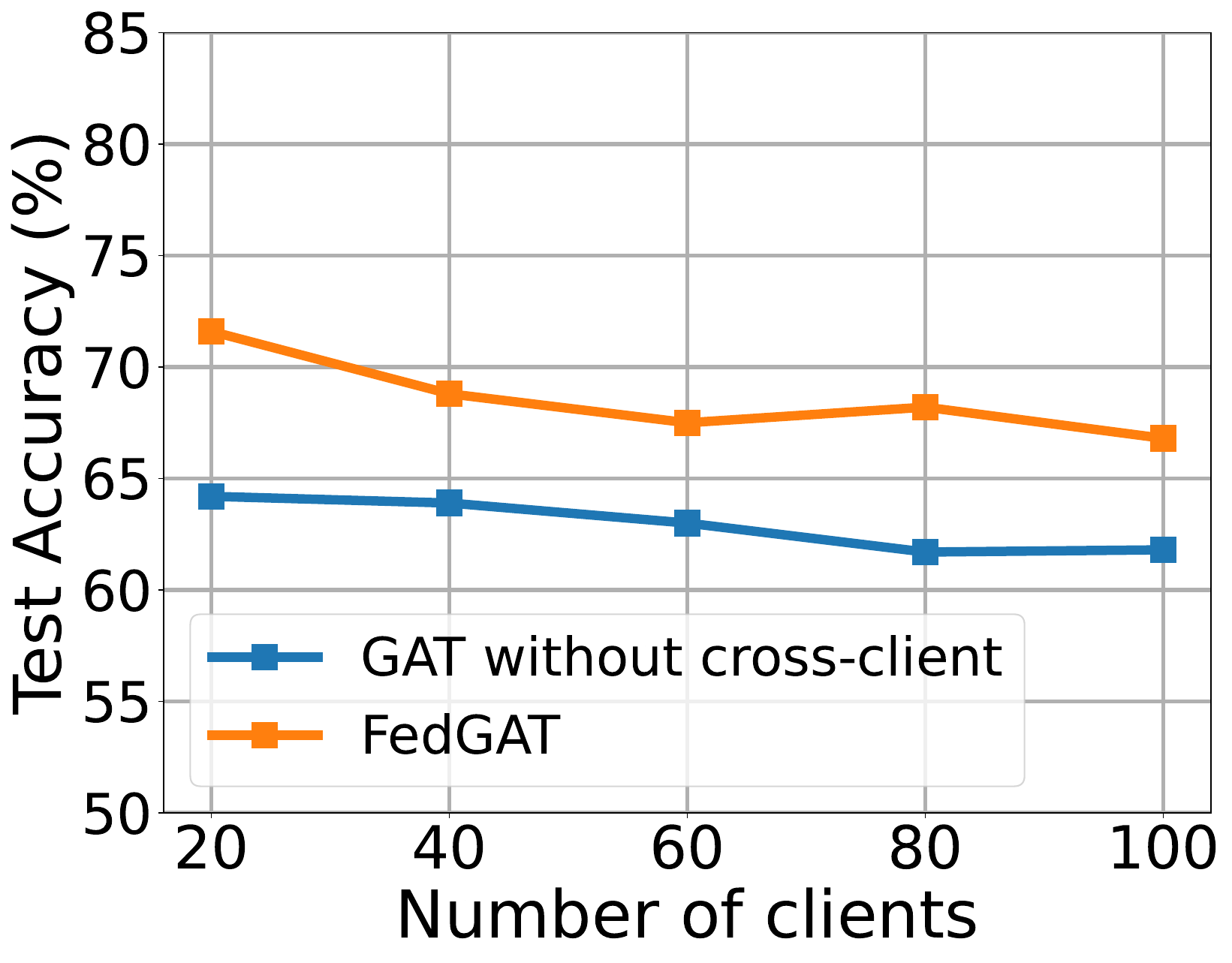}
        \caption{$\beta = 10000$ (iid)}
    \end{subfigure}
    \caption{Accuracy v/s number of clients for iid and non-iid data distribution on \textbf{Cora}(a, b) and \textbf{Citeseer}(c, d) with \textbf{Vector FedGAT}.}
    \label{fig:6}
\end{figure*}

We use more or less the same idea in this algorithm. A polynomial approximation is to be used to compute the attention scores. This is achieved by a computation very similar to the FedGAT algorithm. However, the key difference lies in the elements used in the computation.

The algorithm is motivated by the following observation:

If x is a $1$-dimensional vector, then, element-wise application of the exponentiation operation will yield a vector $x^{n} = [x_{1}^{n}, x_{2}^{n}, ..., x_{d}^{n}]$.

Consider node $i$ in the graph. Let $h_{i}$ denote its node feature vector. Suppose, we are given the expression

$$D = \sum_{j\in\mathcal{N}_{i}}h_{j}u_{j}^{T}$$

where $\{u_{j}\}$ is a set of vectors such that if $u_{j}\in\mathbf{R}^{g}$, then, $\forall n \leq g$, if $u_{j}(n) \neq 0$, then, $u_{k}(n) = 0$ $\forall j\in\mathcal{N}_{j}$. In addition to this, we also assume that the vectors are filled with either $0$'s or $1$'s. We also assume that the dimension of these vectors is strictly larger than the size of the neighbourhood of the node $i$. This is to ensure that the span of the vectors is strictly less than their dimension. For simplicity, we will assume that the dimension of the vectors for node $i$ is $2deg(i)$.

Now, we consider the expressions

$$M_{1i} = mask_{1i} + h_{i}\sum_{j\in\mathcal{N}_{i}}u_{j}^{T}$$

$$M_{2i} = mask_{2i} + \sum_{j\in\mathcal{N}_{i}}h_{j}u_{j}^{T}$$

$$K_{1i} = mask_{3i} + \sum_{j\in\mathcal{N}_{i}}u_{j}h_{j}^{T}$$

$$K_{2i} = mask_{4i}$$

$$K_{3i} = mask_{5i} + \sum_{j\in\mathcal{N}_{i}}u_{j}$$

$mask_{1}, mask_{2}, mask_{3}$ are constructed in such a way that

$$mask_{1i}mask_{4i} = 0$$

$$mask_{2i}mask_{4i} = 0$$

$$\forall u_{j} \text{ , } u_{4}^{T}mask_{3i} = 0$$

$$\forall u_{j} \text{ , } u_{4}^{T}mask_{4i} = u_{j}^{T}$$

Additionally, we also have the constraint that $mask_{5i}$ is a vector orthonormal to all the $u_{j}$'s.

$mask_{1i}, mask_{2i}$ are used to obfuscate the features, since computing outer products of the features with binary vectors will easily reveal any and all node feature information. The use of $mask_{3i}$ and $mask_{4i}$ shall soon become clear. It is trivial to construct such masks.

The way we use these expressions is as follows:

Equations \eqref{eq:4} and \eqref{eq:5}, show a polynomial expansion of the attention scores. If we use a truncated series of maximum degree $p$, as in Equation \eqref{eq:6}, we can compute it as follows using the matrices given above:

\textbf{Step 1: } From Equation \eqref{eq:6}, the expansion of the attention coefficient as a truncated Chebyshev series is

    $$e_{ij} \approx \sum_{n = 0}^{p}q_{n}x_{ij}^{n}$$

    where $x_{ij}$ is defined in Equation \eqref{eq:4}.

    First, we compute $$a_{1}^{T}M_{1i} + a_{2}^{T}M_{2i}$$

    This can be written as

    $$\begin{aligned}
        D_{i} &= a_{1}^{T}M_{1i} + a_{2}^{T}M_{2i}\\
        &= a_{1}^{T}mask_{1i} + a_{2}^{T}mask_{2i} \\
        &+ \sum_{j\in\mathcal{N}_{j}}\bigg(a_{1}^{T}h_{i} + a_{2}^{T}h_{j}\bigg)u_{j}^{T}\\
        &= a_{1}^{T}mask_{1i} + a_{2}^{T}mask_{2i} + \sum_{j\in\mathcal{N}_{i}}x_{ij}u_{j}^{T}\\
    \end{aligned}$$

\textbf{Step 2: } Having computed vector $D_{i}$, we must now get rid of the $mask_{1i}$ term. To do this, we use the vector $K_{2i} = mask_{4i}$. As defined before, $mask_{1i}mask_{4i} = 0$, and $mask_{2i}mask_{4i} = 0$. Also, all the vectors $u_{j}$ 1-eigenvectors of $mask_{4i}$.
    
    Thus, compute

    $$R_{i} = D_{i}mask_{4i}$$

    $$R_{i} = \sum_{j\in\mathcal{N}_{i}}x_{ij}u_{j}^{T}$$

\textbf{Step 3: } Observe an interesting property of $R_{i}$; $R_{i}^{n} = \sum_{j\in\mathcal{N}_{i}}x_{ij}^{n}u_{j}^{T}$. This is because $u_{j}$'s are binary vectors with no overlapping $1$'s. As a result, computing elementwise powers of $R_{i}$ keeps the terms separate throughout.

\textbf{Step 4: } We finally have all the terms we need. We now compute the terms from Equations \eqref{eq:8}, that is, $E_{i}^{(n)}$ and $F_{i}^{(n)}$. We compute it as follows:

    $$E_{i}^{(n)} = R_{i}^{n}K_{1i}$$

    $$F_{i}^{(n)} = R_{i}^{n}K_{2i}$$

    These can then be assembled to compute the GAT update as in Equation \eqref{eq:7}.

\subsection{Communication Complexity of Vector FedGAT}

From the previous section, the terms to be communicated to node $i$ are matrices $M_{1i}$, $M_{2i}$, $K_{1i}$ and $K_{2i}$, $K_{3i}$. The dimensions of $M_{1i}, M_{2i}, K_{1i}$ are $2deg(i)\times d$, where $d$ is the dimension of the feature vectors. This is because we assumed the vectors $u_{j}$ to have dimension equal to $2deg(i)$ for node $i$.

\begin{figure}[t]
    \centering
        \centering
        \includegraphics[width = 0.5\linewidth]{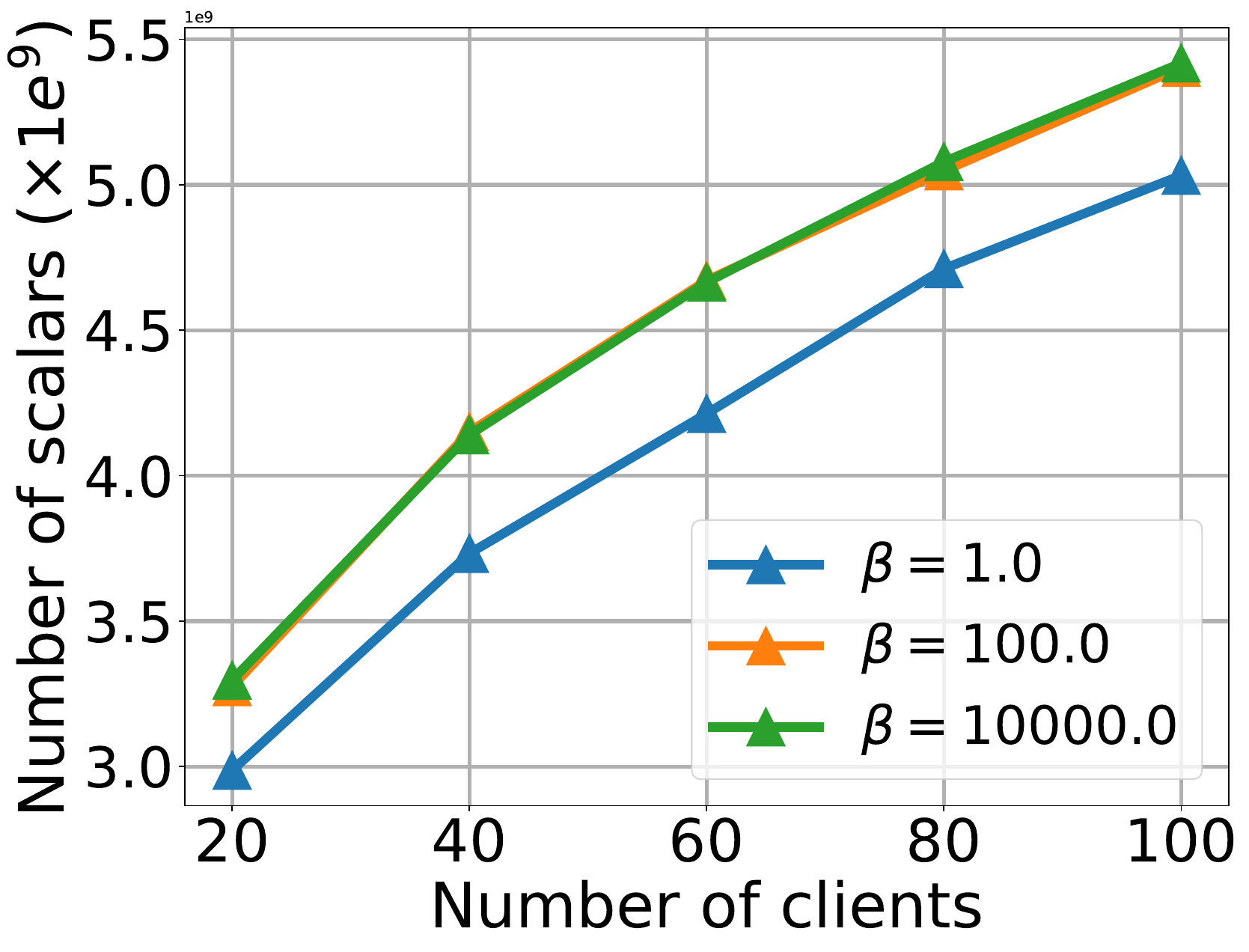}
        \caption{Communication cost for Vector FedGAT on Cora.}
        \label{fig:7}
\end{figure}

\begin{figure}[t]
    \centering
        \centering
        \includegraphics[width = 0.5\linewidth]{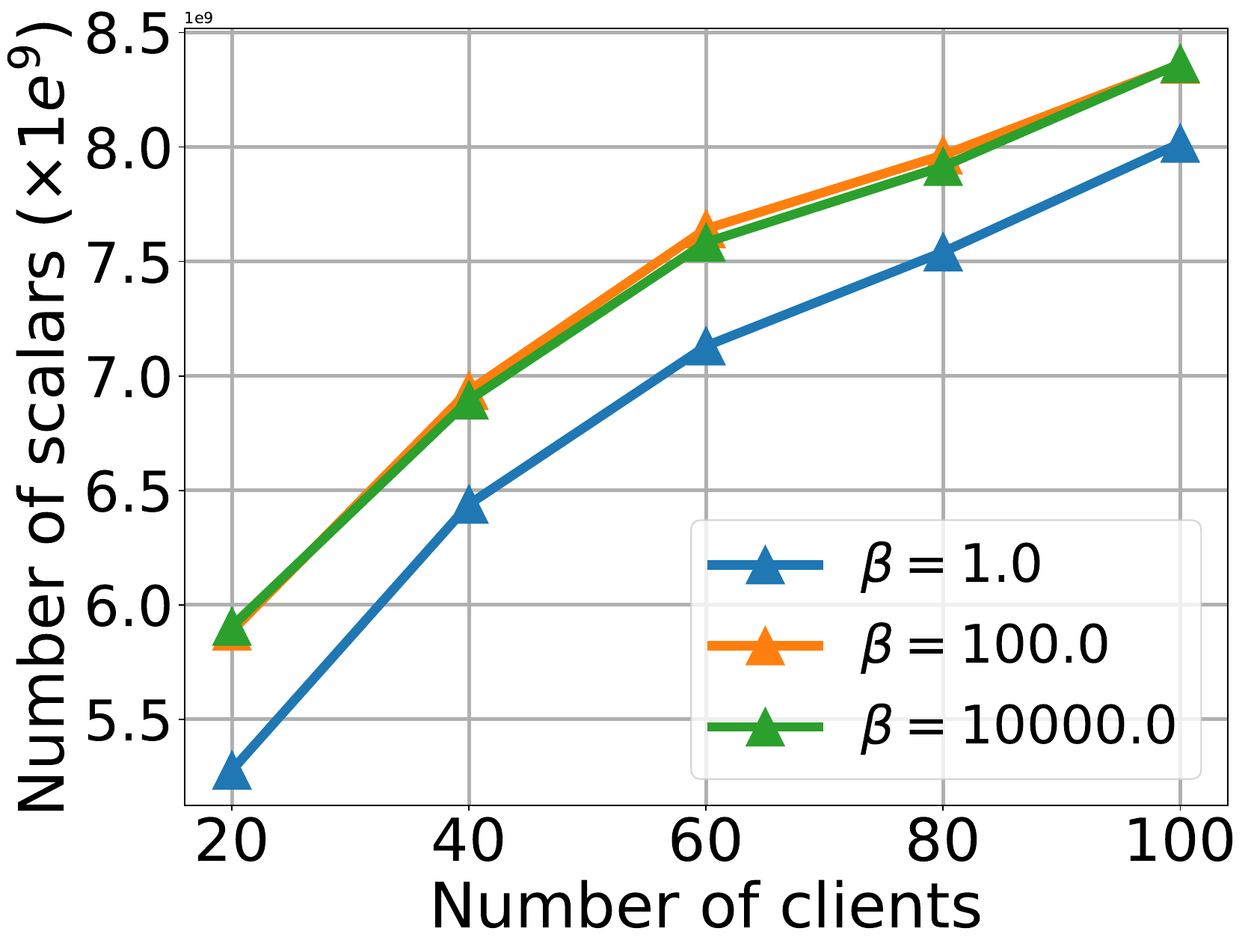}
        \caption{Communication cost for Vector FedGAT on Citeseer.}
        \label{fig:8}
\end{figure}

This term is the largest in the communication for node $i$. Thus, the same proof as FedGAT algorithm can be used to compute the communication cost of the Vector FedGAT; it is simply $\mathcal{O}(KB_{L}dB^{2})$, where the same notation as before is used. This is an improvement over the Matrix version of FedGAT, where the $\mathcal{O}(B^{3})$ dependence was a major bottleneck. The experimental results for the communication overhead of Vector FedGAT are shown in Figures \eqref{fig:7} and \eqref{fig:8}.

\subsection{Experiments for Vector FedGAT}

We performed the same set of basic experiments that we did in the Matrix FedGAT. Here, we show some of the results:

We conducted experiments for the communication overhead of the Vector FedGAT algorithm. We observed a nearly $10\times$ speed-up when compared to the Matrix FedGAT.

As can be seen from Figure \eqref{fig:6}, the performance of the FedGAT algorithm always exceeds the Distributed GAT accuracy. In fact, the difference is even more stark than it was with the number of clients less than 10, as done with Matrix FedGAT. It would have been expensive to do the experiments on a large number of clients with the Matrix version due to the communication overhead. There is no drop in performance as compared to the Matrix version of FedGAT, indicating that the algorithm is sound.

\end{document}